\documentclass{article} 
\usepackage[usenames,dvipsnames,table]{xcolor}
\usepackage{iclr2025_conference,times}

\usepackage{amsmath,amsfonts,bm}

\def\eqref#1{equation~\ref{#1}}

\def\1{\bm{1}}

\DeclareMathAlphabet{\mathsfit}{\encodingdefault}{\sfdefault}{m}{sl}
\SetMathAlphabet{\mathsfit}{bold}{\encodingdefault}{\sfdefault}{bx}{n}

\usepackage{hyperref}
\usepackage{url}
\usepackage{tabularx}
\usepackage{xparse}
\usepackage{seqsplit}
\usepackage{amsmath}
\usepackage{graphicx}
\usepackage{float}
\usepackage{booktabs}
\usepackage{ragged2e}
\usepackage{longtable}
\usepackage{pgfplots}
\usepackage{tabularx}
\usepackage{array}
\usepackage{subcaption}
\usepackage{multirow}
\usepackage{amssymb}
\usepackage{adjustbox}
\usepackage{makecell}
\usepackage{placeins}
\usepackage{makecell}

\ExplSyntaxOn
\NewDocumentCommand{\authorlist}{m}
 {
  \seq_set_split:Nnn \l_tmpa_seq { , } { #1 }
  \seq_map_indexed_function:NN \l_tmpa_seq \__author_wrap:nn
 }

\cs_new_protected:Nn \__author_wrap:nn
 {
  \mbox{#2}%
  \bool_if:nTF { \int_compare_p:nNn {#1} < { \seq_count:N \l_tmpa_seq } }
    {, \space}
    {}
 }
\ExplSyntaxOff

\title{Dynamic-SUPERB Phase-2: A Collaboratively Expanding Benchmark for Measuring the Capabilities of Spoken Language Models with 180 Tasks}

\author{%
  \parbox[t]{\linewidth}{%
    \raggedright\authorlist{Chien-yu Huang$^{1}$, Wei-Chih Chen$^{1}$, Shu-wen Yang$^{1}$, Andy T. Liu$^{1}$, Chen-An Li$^{1}$, Yu-Xiang Lin$^{1}$, Wei-Cheng Tseng$^{2}$, Anuj Diwan$^{2}$, Yi-Jen Shih$^{2}$, Jiatong Shi$^{3}$, William Chen$^{3}$, Chih-Kai Yang$^{1}$, Wenze Ren$^{1}$, Xuanjun Chen$^{1}$, Chi-Yuan Hsiao$^{1}$, Puyuan Peng$^{2}$, Shih-Heng Wang$^{1}$, Chun-Yi Kuan$^{1}$, Ke-Han Lu$^{1}$, Kai-Wei Chang$^{1}$, Fabian Ritter-Gutierrez$^{4}$, Kuan-Po Huang$^{1}$, Siddhant Arora$^{3}$, You-Kuan Lin$^{1}$, Ming To Chuang$^{1}$, Eunjung Yeo$^{3}$, Kalvin Chang$^{3}$, Chung-Ming Chien$^{5}$, Kwanghee Choi$^{3}$, Jun-You Wang$^{6}$, Cheng-Hsiu Hsieh$^{1}$, Yi-Cheng Lin$^{1}$, Chee-En Yu$^{1}$, I-Hsiang Chiu$^{1}$, Heitor R. Guimarães$^{7}$, Jionghao Han$^{3}$, Tzu-Quan Lin$^{1}$, Tzu-Yuan Lin$^{8}$, Homu Chang$^{1}$, Ting-Wu Chang$^{1}$, Chun Wei Chen$^{1}$, Shou-Jen Chen$^{1}$, Yu-Hua Chen$^{1}$, Hsi-Chun Cheng$^{1}$, Kunal Dhawan$^{9}$, Jia-Lin Fang$^{1}$, Shi-Xin Fang$^{1}$, Kuan-Yu Fang Chiang$^{1}$, Chi An Fu$^{1}$, Hsien-Fu Hsiao$^{1}$, Ching Yu Hsu$^{1}$, Shao-Syuan Huang$^{1}$, Lee Chen Wei$^{1}$, Hsi-Che Lin$^{1}$, Hsuan-Hao Lin$^{1}$, Hsuan-Ting Lin$^{1}$, Jian-Ren Lin$^{1}$, Ting-Chun Liu$^{1}$, Li-Chun Lu$^{1}$, Tsung-Min Pai$^{1}$, Ankita Pasad$^{5}$, Shih-Yun Shan Kuan$^{1}$, Suwon Shon$^{10}$, Yuxun Tang$^{11}$, Yun-Shao Tsai$^{1}$, Jui-Chiang Wei$^{1}$, Tzu-Chieh Wei$^{1}$, Chengxi Wu$^{1}$, Dien-Ruei Wu$^{1}$, Chao-Han Huck Yang$^{9}$, Chieh-Chi Yang$^{1}$, Jia Qi Yip$^{4}$, Shao-Xiang Yuan$^{1}$, Vahid Noroozi$^{9}$, Zhehuai Chen$^{9}$, Haibin Wu$^{1}$, Karen Livescu$^{5}$, David Harwath$^{2}$, Shinji Watanabe$^{3}$, Hung-yi Lee$^{1}$}%
  }\\[10ex]
  \\
  $^{1}$National Taiwan University \quad $^{2}$University of Texas at Austin \quad $^{3}$Carnegie Mellon University \\
  $^{4}$Nanyang Technological University \quad $^{5}$Toyota Technological Institute at Chicago \\
  $^{6}$Academia Sinica \quad $^{7}$Université du Québec (INRS-EMT) \quad $^{8}$Independent Researcher \\
  $^{9}$NVIDIA \quad $^{10}$ASAPP \quad $^{11}$Renmin University of China
}

\iclrfinalcopy 
\begin{document}

\maketitle

\begin{abstract}
Multimodal foundation models, such as Gemini and ChatGPT, have revolutionized human-machine interactions by seamlessly integrating various forms of data.
Developing a universal spoken language model that comprehends a wide range of natural language instructions is critical for bridging communication gaps and facilitating more intuitive interactions.
However, the absence of a comprehensive evaluation benchmark poses a significant challenge.
We present Dynamic-SUPERB \textit{Phase-2}, an open and evolving benchmark for the comprehensive evaluation of instruction-based universal speech models.
Building upon the first generation, this second version incorporates 125 new tasks contributed collaboratively by the global research community, expanding the benchmark to a total of \textit{180 tasks}, making it the largest benchmark for speech and audio evaluation.
While the first generation of Dynamic-SUPERB was limited to classification tasks, Dynamic-SUPERB \textit{Phase-2} broadens its evaluation capabilities by introducing a wide array of novel and diverse tasks, including regression and sequence generation, across speech, music, and environmental audio.
Evaluation results show that no model performed well universally.
SALMONN-13B excelled in English ASR and Qwen2-Audio-7B-Instruct showed high accuracy in emotion recognition, but current models still require further innovations to handle a broader range of tasks.
We open-source all task data and the evaluation pipeline at \url{https://github.com/dynamic-superb/dynamic-superb}.
\end{abstract}

\section{Introduction}
Recent advancements in large language models (LLMs) have accelerated the development of natural language processing (NLP) \citep{touvron2023llama, achiam2023gpt, li2023textbooks, claude, bai2023qwen}.
These models can follow natural language instructions, making users quickly adopt them for a variety of applications.
They have been integrated into commercial products, such as ChatGPT \citep{achiam2023gpt} and Claude \citep{claude}, as well as in the open-source research community, including the LLaMA series \citep{touvron2023llama, touvron2023llama2, dubey2024llama}.
Yet, they are primarily text-based models, meaning they cannot process speech or audio, which are essential for more natural ways of communication and interaction with the real world.

Compared to written text, spoken language has always been a more natural and convenient way for humans to communicate.
Spoken language conveys a wealth of information, including semantics, prosody, emotion, and speaker characteristics, while text is limited to representing semantic information, which can sometimes even depend on the prosodic cues present in spoken language \citep{lin2024paralinguistics}.
This highlights the need for universal speech models and explains why automatic speech recognition (ASR) systems using text-based language models are not optimal.
Several attempts have been made to develop instruction-based universal speech or audio models capable of performing various tasks, such as LTU-AS \citep{gong2023joint}, SALMONN \citep{tang2024salmonn}, Qwen-Audio \citep{chu2023qwen, chu2024qwen2}, and WavLLM \citep{hu2024wavllm}.

Despite significant research in universal speech models, evaluating them effectively and comprehensively remains a major challenge.
In NLP, benchmarks for text LLMs include a large number of tasks.
CrossFit \citep{ye2021crossfit} includes 160 tasks, BIG-bench \citep{srivastava2022beyond} comprises 204 tasks, and Natural-Instructions \citep{naturalinstructions, supernaturalinstructions} provides 1,616 tasks.
For universal speech models, several benchmarks have been proposed but with a \textit{fixed} and \textit{limited} set of tasks (typically around a dozen) to evaluate specific capabilities of models \citep{yang2021superb, dunbar2022self, maimon2024salmon}.
This is insufficient for evaluating a universal model, as we aim to investigate whether models can handle a broader range of tasks beyond fixed benchmarks.
There is a strong demand for a benchmark that can evaluate these models across various aspects with a wide range of speech tasks, which led to the creation of Dynamic-SUPERB \citep{huang2024dynamic}.

Dynamic-SUPERB is the first benchmark for evaluating universal instruction-based speech and audio models.
As models advance, we \textit{dynamically} expand the benchmark by adding new tasks to provide better guidance for researchers in model development.
We have designed a well-structured pipeline that harnesses the collective efforts of the community to gather diverse challenging, novel, and creative tasks.
The first generation of Dynamic-SUPERB consists of 55 tasks, covering various aspects of speech (e.g., semantics, speakers, etc.), making it the speech benchmark with the most tasks.
However, this is not sufficient to pave the way for developing universal speech models, especially given the complexity and richness of the information conveyed by spoken language.

This paper presents the \textbf{Dynamic-SUPERB Phase-2}.
We expanded it to \textit{180 tasks}, more than double the size of its first iteration, with contributions from the global research community.
They cover a broad range of types, and some tasks present novel challenges that have not been explored in any previous research.
Besides, previous speech research has typically treated music and environmental audio as background noise to be ignored.
However, these sounds contain rich information and share overlapping elements that complement each other.
Thus, in Dynamic-SUPERB, we also introduced preliminary music and audio tasks from established music and audio benchmarks \citep{yuan2023marble, turian2022hear}.
We define the core tasks by reformulating the SUPERB \citep{yang2021superb} (speech), MARBLE \citep{yuan2023marble} (music), and HEAR \citep{turian2022hear} (audio) benchmarks for quick-round research experiments.
One challenge of evaluating with such a large-scale benchmark is deriving concrete and useful insights from hundreds of evaluation results.
We provide a taxonomy for every task in Dynamic-SUPERB, where tasks are clustered by the specific model capabilities they probe.
Researchers can follow this taxonomy to develop or reinforce specific capabilities of the models they build.
To our knowledge, Dynamic-SUPERB is the largest benchmark and the first to provide such a detailed task taxonomy in speech processing.

We conducted a comprehensive evaluation of several models using Dynamic-SUPERB \textit{Phase-2}.
To handle the diverse output formats of these models, we propose an automated LLM-based pipeline for general evaluation across tasks.
The evaluation results show that these models perform well only on a limited range of tasks.
For example, SALMONN-13B performed well on English ASR, while Qwen2-Audio-7B-Instruct achieved high accuracy in emotion recognition.
However, they do not generalize well to a much broader range of tasks in speech recognition and paralinguistics.
Notably, training on diverse data, even with significant differences in signal-level characteristics, can enhance performance across domains.
We observed that spoken language models outperformed music models in certain music tasks.
This highlights the potential for developing unified models for speech, music, and general audio.
We have open-sourced all materials to support reproducibility and further research.
We invite researchers to help expand Dynamic-SUPERB, making it more diverse and comprehensive to advance universal spoken language models.

\section{Related works}
\subsection{Instruction-following universal speech models}
LLMs have shown strong natural language processing abilities and are used in speech, audio, and music applications.
Recent frameworks integrate a pre-trained speech encoder with an LLM through fine-tuning techniques such as LoRA \citep{hu2021lora}.
LTU-AS \citep{gong2023joint} combines Whisper \citep{radford2023robust} with LLaMA \citep{touvron2023llama}, and it is fine-tuned on open-ended speech and audio question-answering.
SALMONN adopts a window-level Q-former \citep{li2023blip} to generate soft prompts fusing representations from the speech and audio encoders.
Qwen-Audio \citep{chu2023qwen} introduces task-specific tags into Qwen to encourage knowledge sharing and prevent interference across diverse tasks, and it supports multi-turn dialogues for both audio and text inputs.
WavLLM \citep{hu2024wavllm} uses curriculum learning to prevent overfitting on specific tasks while maintaining the LLM's original capabilities.
All these models accept speech, audio, and text as input but output only text, focusing on \textit{understanding} rather than \textit{generation} in speech and audio.
Several attempts have also used prompts for speech, music, and audio generation \citep{guo2023prompttts, agostinelli2023musiclm, liu2023audioldm}.
However, to our knowledge, there is no universal model capable of handling both generation and understanding tasks.
Moreover, these models have not been comprehensively evaluated on a benchmark, which hinders fair comparisons between them.

\subsection{Evaluation benchmarks}
Several benchmarks have been developed to evaluate speech models.
SUPERB \citep{yang2021superb} is the most widely used benchmark for assessing the performance of speech foundation models across various tasks that cover different aspects of speech.
On the other hand, SLUE \citep{shon2022slue, shon2023slue} focuses more specifically on spoken language understanding.
Yet, they are primarily limited to English.
LeBenchmark \citep{evain2021lebenchmark, parcollet2023lebenchmark} evaluates speech foundation models specifically for French, while IndicSUPERB \citep{javed2023indicsuperb} is dedicated to Indian languages.
ML-SUPERB \citep{shi2023mlsuperb, shi2024mlsuperb} offers ASR and language identification tasks in 100+ languages.
Aside from speech, MARBLE \citep{yuan2023marble} and HEAR \citep{turian2022hear} offer platforms for evaluating various music and a wide range of audio tasks.
Using these benchmarks, researchers build a specialized model for each task.
Thus, the number of tasks is limited due to the growing costs associated with adding more tasks.
Conversely, universal models are expected to do various tasks without fine-tuning for each one, allowing users to engage in far more diverse applications and enabling benchmark developers to include more tasks for comprehensive evaluation.
In NLP and CV, benchmarks have been developed to evaluate models across a much broader range of tasks \citep{bitton2023visit, srivastava2022beyond, naturalinstructions}.
However, in speech, we lack a benchmark of comparable scale.
Dynamic-SUPERB \citep{huang2024dynamic} is the first benchmark for instruction-based universal speech models.
While its first version includes far more tasks than all other speech benchmarks, they are all classification tasks.
AIR-bench \citep{yang2024air}, on the other hand, includes tasks from speech, music, and audio, and expands beyond classification, although the number of tasks available for evaluating universal models remains limited (19 tasks in foundation track).
Thus, there remains a critical need for a comprehensive benchmark that evaluates universal models across a broader range of tasks to fully assess their capabilities.
Consequently, we developed Dynamic-SUPERB \textit{Phase-2}, substantially upgrading the first generation with a detailed taxonomy and establishing community contribution protocols to facilitate continual task integration.

\section{Dynamic-SUPERB Phase-2}

\begin{figure}[ht]
    \centering
    \includegraphics[width=0.8\linewidth]{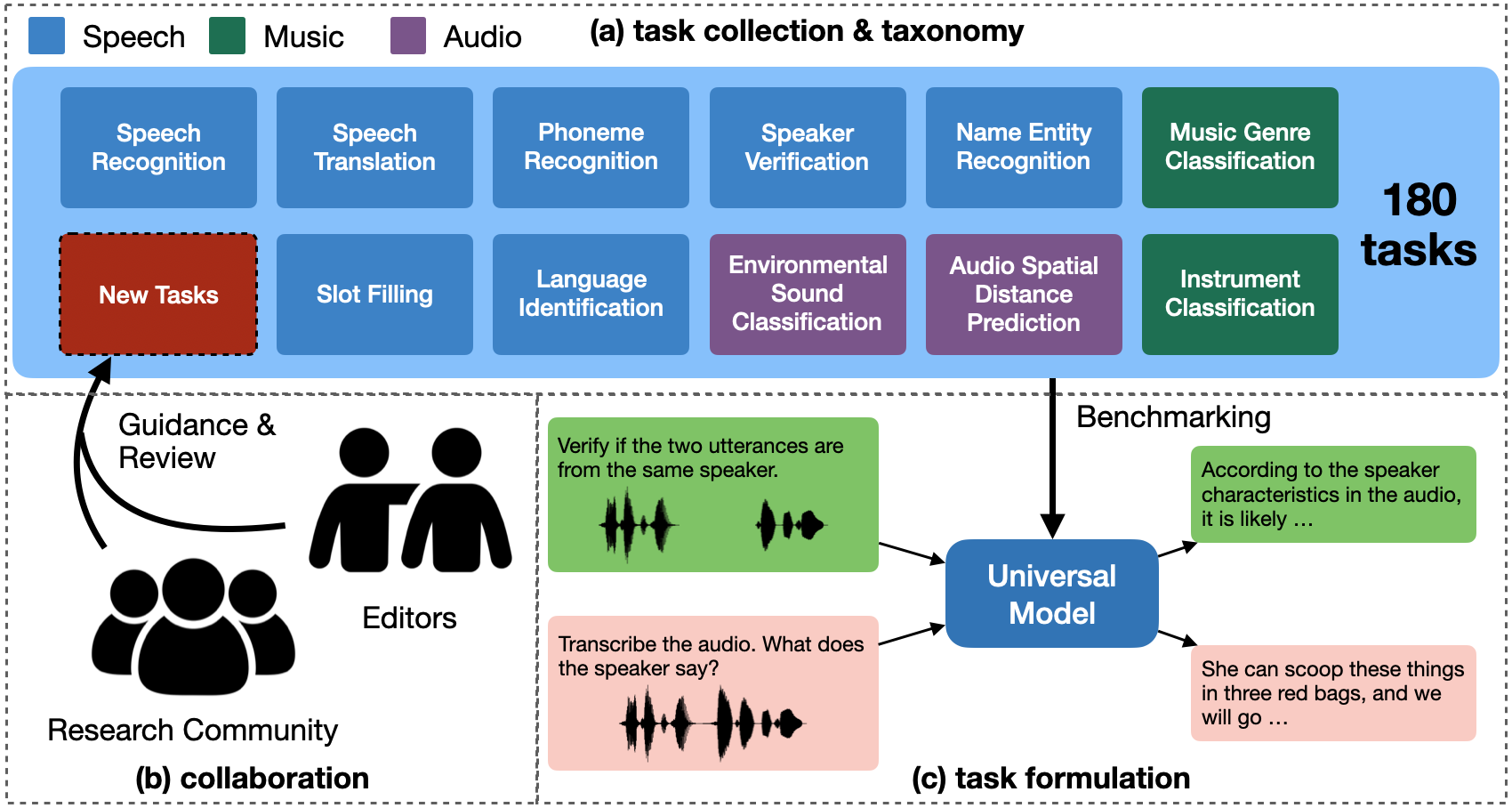}
    \caption{An overview of Dynamic-SUPERB.}
    \label{fig:ds-overview}
    \vspace{-1em}
\end{figure}

\subsection{Overview}
Figure~\ref{fig:ds-overview} depicts the framework of Dynamic-SUPERB \textit{Phase-2}.
Our goal is to evaluate universal models that meet the following criteria:
(1) The model can accept speech, music, or other audio as input.
(2) The model can perform specific tasks without requiring fine-tuning.
(3) The model follows natural language instructions to execute the corresponding tasks.
To comprehensively evaluate these universal models, we have collected hundreds of tasks and developed a task taxonomy to better guide benchmark users (Figure~\ref{fig:ds-overview}a).
All tasks in Dynamic-SUPERB \textit{Phase-2} are intended solely for testing purposes; we do not provide training data for two main reasons.
First, current models are trained on large-scale, open-source, or proprietary datasets, making it challenging to ensure that all models are trained under consistent conditions.
Second, we aim to evaluate universal models that do not require fine-tuning for downstream tasks.

The Dynamic-SUPERB project is designed to evolve dynamically with research advancements by incorporating novel tasks from the research community (Figure~\ref{fig:ds-overview}b).
With the first call for tasks, Dynamic-SUPERB \textit{Phase-2} has grown from its first generation, expanding from 55 to 180 tasks.
Table~\ref{tab:benchmark-comparison} compares several benchmarks used in speech, music, and audio research, demonstrating that Dynamic-SUPERB \textit{Phase-2} is the largest benchmark covering all three areas.
It provides a more fine-grained evaluation than any other benchmark.

\begin{table}[ht]
    \centering
    \caption{Comparison of popular benchmarks with Dynamic-SUPERB.}
    \label{tab:benchmark-comparison}
    \footnotesize
    \setlength{\tabcolsep}{4pt}
    \resizebox{.95\linewidth}{!}{
    \begin{tabularx}{\linewidth}{c c c c c c c c}
         \toprule
         \multirow{2}{*}{\textbf{Benchmark}} & \multirow{2}{*}{SUPERB} & \multirow{2}{*}{SLUE} & \multirow{2}{*}{HEAR} & \multirow{2}{*}{MARBLE} & \multirow{2}{*}{AIR-Bench} & {\footnotesize Dynamic-} & {\footnotesize Dynamic-} \\
         {} & {} & {} & {} & {} & {} & {\footnotesize SUPERB Phase-1} & {\footnotesize SUPERB \textit{Phase-2}}\\
         \midrule
         \textbf{\# of tasks} & 13 & 7 & 19 & 13 & 19$^*$ & \textbf{55} & \textbf{180} \\
         \midrule
         \textbf{Speech} & \checkmark & \checkmark & {} & {} & \checkmark & \checkmark & \checkmark \\
         \textbf{Music} & {} & {} & {} & \checkmark & \checkmark & \checkmark & \checkmark \\
         \textbf{Audio} & {} & {} & \checkmark & {} & \checkmark & \checkmark & \checkmark \\
         \bottomrule
    \end{tabularx}
    }
    \justify
    \footnotesize{$^*$ foundation track}
    \vspace{-1.2em}
\end{table}

\subsection{Task formulation}
As Figure~\ref{fig:ds-overview}c depicts, in Dynamic-SUPERB \textit{Phase-2}, each task is structured to include:
(1) Text instruction: A natural language instruction that guides the model on the task to perform.
Each task has multiple different instructions to evaluate the model's ability to understand instructions.
(2) Audio component: At least one audio element, which can be in the input, the output, or both.
(3) (Optional) Text component: Text elements (other than instruction) that may serve as inputs or outputs.
The number of audio elements in the inputs or outputs may vary depending on the task.
For example, in speaker verification, two utterances are used to determine whether they were produced by the same speaker (green blocks in Figure~\ref{fig:ds-overview}c).
Besides, text inputs are not always required; for instance, ASR does not involve any text inputs.
We use text format for instructions instead of spoken format.
Spoken instructions involve varying content, speaker characteristics, and prosody, making them more complex.
Text instructions act as an intermediary, bridging text and spoken universal models.
These designs ensure consistency and simplify the model’s understanding and processing of diverse tasks while maintaining the benchmark's extendibility with minimal constraints.

Another problem in evaluating universal models is the format of classification and regression tasks.
Generally, a task-specialized model generates predicted labels (soft distributions or hard labels) within a pre-defined set of labels for classification or numeric values for regression within a pre-defined range.
However, a universal model does not necessarily have access to this information, and thus it is hard to evaluate them in the same way, especially using it to perform several different tasks.
To address this, we believe that a universal speech model should produce outputs in natural language for all tasks (model outputs in Figure~\ref{fig:ds-overview}c).
Therefore, in Dynamic-SUPERB, \textit{all classification labels are represented as text.}
For regression tasks with specific formats (e.g., scalars, JSON, or Python-style lists), we can parse the natural language outputs using a post-processing pipeline (Section~\ref{sec:evaluation-metrics}).

\subsection{Call for tasks}
Dynamic-SUPERB fosters community collaboration, encouraging the addition of innovative tasks to remain relevant in this rapidly changing field.
In \textit{Phase-2}, we initiated a call for tasks in March 2024 to invite contributions from the research community.
We established an organized and transparent submission process on our GitHub portal, where we organize members who serve as editors to guide contributors.
Contributors propose tasks by providing relevant information on GitHub.
Each proposal is assigned to an editor who checks for major issues and prevents duplicated efforts.
Then, task proposers upload their data to our Huggingface space in specified formats, using only datasets with proper licenses.
They complete submissions by opening a pull request on GitHub with the necessary files.
Afterward, editors review submissions on a rolling basis, offering suggestions for refinement rather than immediate acceptance or rejection.
After iterative improvements, accepted tasks are merged into the repository.
Between March and July 2024, we received more than 140 task proposals and accepted over 120 new tasks.
Tasks still under review are not included here but will be featured in the next phase.
For details, please refer to Appendix~\ref{sec:appendix-call-for-tasks}.

\subsection{Task taxonomy}

\begin{figure}[h]
    \centering
    \begin{subfigure}{.47\linewidth}
        \centering
        \includegraphics[width=\linewidth]{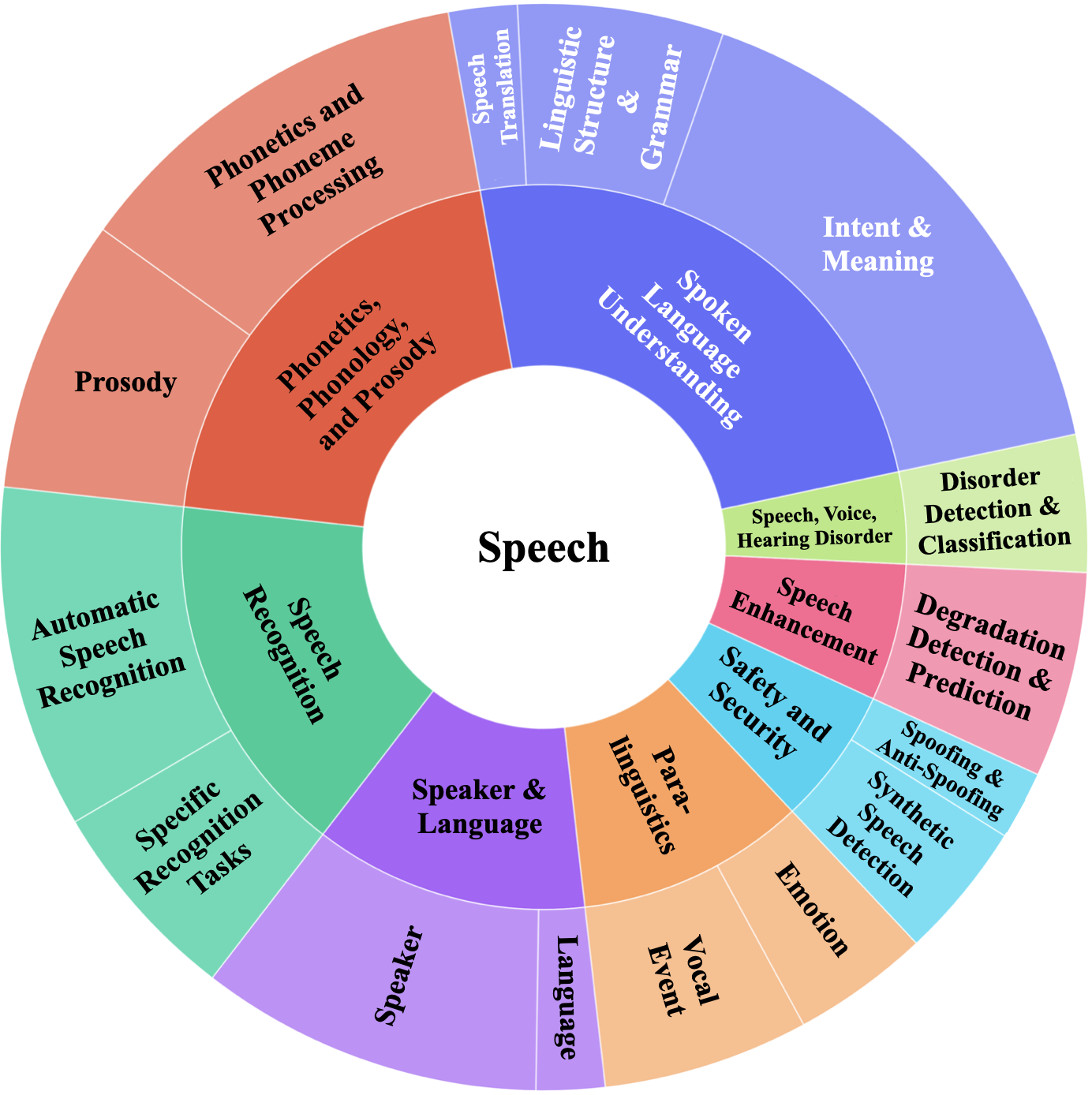}
        \caption{Task Taxonomy for speech tasks.}
        \label{fig:task-taxonomy-speech}
    \end{subfigure}
    \hfill
    \begin{subfigure}{.47\linewidth}
        \centering
        \includegraphics[width=\linewidth]{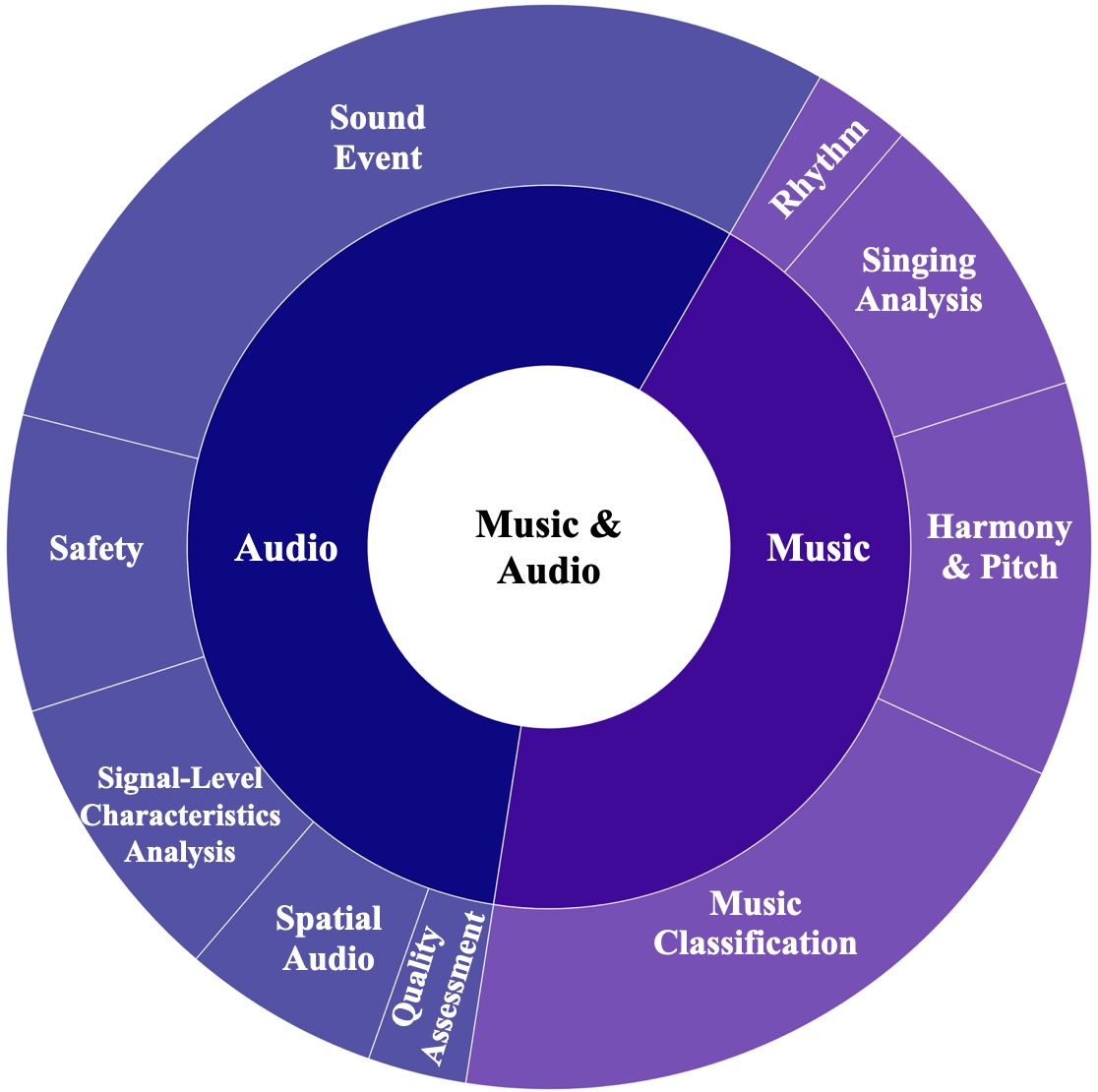}
        \caption{Task Taxonomy for audio and music tasks.}
        \label{fig:task-taxonomy-audio-music}
    \end{subfigure}
    \caption{Task taxonomy in Dynamic-SUPERB.}
    \label{fig:task-taxonomy-meta}
\end{figure}

One primary challenge in building a large-scale benchmark is offering valuable insights to its users.
To address this, we developed a task taxonomy\footnote{We developed the taxonomy by referencing sessions from the INTERSPEECH conference and EDICS of IEEE SPS. Though not identical, they have a very similar structure (please refer to Appendix~\ref{sec:appendix-task-taxonomy} for comparison).} that helps researchers interpret performance results across various tasks.
Researchers can leverage this taxonomy to select specific tasks for model development instead of evaluating every task in the benchmark.

Figures~\ref{fig:task-taxonomy-meta} shows the high-level task taxonomy in Dynamic-SUPERB \textit{Phase-2}.
Due to the space limitation, we only show some representative tasks.
Each leaf node includes at least one task and may be further categorized into more fine-grained sub-domains, which are not shown here due to space constraints.
For example, within `Speaker \& Language/Speaker', we have two categories: `Speaker Characteristics' and `Speaker Identification', each containing several tasks.
Please refer to Appendix~\ref{sec:complete-task-list} for the complete task list.
We first categorize tasks into two primary fields: (I) speech and (II) music \& audio, which are generally distinguished by the source of the sound.
For instance, speech is produced by human vocal cords, music is created using instruments, and audio includes sounds generated by other creatures, materials, or natural phenomena.
We then split each field into several domains based on the key attributes and challenges that the tasks within them present.

\subsubsection{Speech}
As Figure~\ref{fig:task-taxonomy-speech} shows, there are 8 domains within the speech field.
\textbf{Speech Recognition} focuses on converting spoken language into text.
This includes tackling various challenges like multilingual and code-switching ASR, as well as spontaneous ASR which is very different from audiobook-style ones.
It also contains specialized tasks such as command recognition and keyword spotting.
\textbf{Speaker and Language} addresses the analysis of speaker characteristics and languages, covering tasks such as speaker verification, diarization, and language identification.
\textbf{Spoken Language Understanding} deals with understanding and analysis of the content and semantics of spoken language.
It covers tasks like sentiment analysis and speech translation.
\textbf{Phonetics, Phonology, and Prosody} focuses on the sound structure of speech, including phoneme recognition, pronunciation evaluation, and prosodic features like stress and accent classification.
\textbf{Paralinguistics} explores non-verbal aspects of speech, such as emotion recognition and vocal event detection, which can capture nuances like screaming or coughing.
\textbf{Speech Enhancement} aims to improve speech quality by detecting and mitigating noise, reverberation, and other degradations, but currently, we only have understanding tasks.
\textbf{Speech, Voice, and Hearing Disorders} is dedicated to identifying and classifying disorders such as stuttering and so on.
\textbf{Safety and Security} focuses on detecting synthetic or manipulated speech, addressing tasks like spoof detection and recognizing deepfake voices.

\subsubsection{Audio \& Music}
The audio and music field includes a wide range of tasks that focus on various attributes of sound beyond speech, such as musical elements, environmental sounds, and advanced sound analysis.
This field is divided into 9 domains, each addressing a specific aspect of audio or music processing.
\textbf{Music Classification} tasks focus on categorizing musical elements such as instruments, genres, and emotions, providing a foundation for recognizing and analyzing different types of musical content.
\textbf{Pitch Analysis} delves into identifying the pitch and harmony within music, including tasks like pitch extraction, chord classification, and key detection.
\textbf{Rhythm Analysis} involves tasks such as beat tracking, which is critical for understanding the temporal structure of music.
In \textbf{Singing Analysis}, tasks address both lyric recognition and translation, as well as the classification of vocal techniques used in singing.
\textbf{Quality Assessment} evaluates the perceived quality of singing, including automated predictions of Mean Opinion Scores (MOS).
The \textbf{Sound Event} domain is broader, covering various sound sources such as animals, the environment, and human activities. Tasks range from animal sound classification to emergency traffic detection and even advanced tasks like multi-channel sound event understanding.
\textbf{Safety} domain includes detecting deepfakes in singing voices and identifying manipulated audio files, ensuring sound authenticity and integrity.
\textbf{Spatial Audio} covers all tasks related to understanding spatial information, such as estimating the distance or position of sounds in the real world.
Finally, \textbf{Signal Characteristics Analysis} domain addresses general signal-level characteristics of audio, including sound effect detection, and duration prediction.

\subsubsection{Core tasks}

While our task taxonomy provides a comprehensive, hierarchical, and systematic approach for benchmark users to easily get started with Dynamic-SUPERB \textit{Phase-2}, evaluating all tasks with limited resources remains a challenge.
To address this, alongside the task taxonomy, we have selected \textbf{core tasks} for speech, music, and audio.
These core tasks are reduced subsets of essential tasks that have been widely used or studied within the research community.
We include tasks from three popular benchmarks: SUPERB (speech), MARBLE (music), and HEAR (audio).
The three benchmarks were originally designed for evaluating encoders, not for an instruction-following universal model, so modifications are required.
We crafted instructions for each task, reduced the data size, and reformulated them into the Dynamic-SUPERB format.
Besides, we replaced the datasets in some tasks to address licensing issues.
Since the core task set is much smaller than the full benchmark, researchers can more efficiently evaluate their models across a reasonable range of domains.

\section{Experimental settings}
\subsection{Models}
\label{sec:baselines}

We evaluated several publicly-available models on Dynamic-SUPERB \textit{Phase-2}: SALMONN, LTU-AS, Qwen-Audio-Chat, Qwen2-Audio-7B-Instruct, WavLLM, MU-LLaMA \citep{liu2024music}, GAMA \citep{ghosh2024gama}.
SALMONN is further categorized into 7B and 13B versions based on the size of its LLM component.
All of these models are publicly available, and we utilized their official implementations for inference without any modifications.
The first five models are instruction-based speech models, each also trained with audio understanding capabilities.
However, as they were not trained on music data, we do not expect them to perform well on music-related tasks.
Hence, we further included MU-LLaMA and GAMA.
MU-LLaMA is specifically designed for various music-understanding tasks, and GAMA is trained with both general audio data and a music corpus.
Similar to the speech models, MU-LLaMA and GAMA adopt an LLM-based framework.
They use a music or audio encoder, such as MERT \citep{li2024mert} or Q-former, to convert music or audio into features, which the LLM then uses as prompts for subsequent reasoning.
The sampling strategies used for generating outputs from the LLMs were retained as defined in their official implementations.
Finally, we implemented a cascaded system baseline called Whisper-LLaMA.
The system first transcribes audio using Whisper-v3-large, and then LLaMA3.1-8B processes the transcriptions to perform various tasks based on the provided instructions.

When evaluating models on these diverse tasks, several challenges inevitably arise.
One challenge was testing the baseline models on tasks involving multiple audio inputs.
For example, in speaker verification, a model determines whether two utterances are produced by the same speaker.
Among the evaluated models, only Qwen-Audio and Qwen2-Audio provide interfaces that allow inputting multiple audio files.
For the remaining models, we concatenated all audio files, separated by 0.5 seconds of silence, and used the concatenated file as input.

Last, different models have their own maximum supported audio durations.
Models using Whisper to encode speech face an inherent limitation: Whisper can process audio files of up to 30 seconds, and some models are restricted to handling even shorter durations.
Consequently, we retained the original model settings and did not modify the model architecture to accommodate longer audio clips.
A preliminary analysis of all audio (Appendix~\ref{sec:audio-duration}) in Dynamic-SUPERB \textit{Phase-2} shows that only a small proportion (around 6.57\%) exceeds 30 seconds in length.
Hence, we believe our settings do not largely impact the evaluation results and ensure reasonable inference efficiency.

\subsection{Evaluation metrics}
\label{sec:evaluation-metrics}

The outputs of universal speech models are natural language sentences, making it difficult to assess their correctness using conventional metrics.
For classification tasks, such as emotion recognition, a task-specific model outputs a label from predefined emotions in the dataset, while a universal model generates sentences like ``The speaker sounds happy.''
In this case, we can easily assess the correctness of the former by comparing labels, but this approach does not apply to the latter.
Similarly, in regression tasks, a natural language response like ``Using MOS scoring criteria, I give the audio a score of 3 out of 5'' makes it difficult to directly use metrics such as mean square error.

In evaluation, we categorize tasks into three types: (1) classification, (2) regression, and (3) sequence generation.
For each type, we use different pipelines to evaluate the models' outputs.
For classification tasks, we utilize external LLMs (GPT-4o) as \textit{referees}, with the temperature set to be zero for evaluation consistency, to evaluate whether the outputs from speech or music models\footnote{To avoid ambiguity, unless specified otherwise, the terms ``output'' or ``model output'' in this section refer to the results generated by the models in Sec.~\ref{sec:baselines}.} match the ground truth.
This approach has been widely adopted in the NLP community \citep{wang2023chatgpt, liu2023g, chiang2023can}, and we extend its application to speech research.
We design a prompt that includes the task instructions, the model’s output to be evaluated, and the corresponding ground truth.
The LLM judge gets this prompt and determines if the output aligns with the ground truth using a chain-of-thought reasoning process.
By leveraging the strong instruction-following capabilities of these LLMs, we constrain them to provide the final decision in a fixed format, allowing us to extract the answer using simple methods such as regular expressions\footnote{We have tried to ask the speech universal models to directly output their decisions in a fixed format, but they do not always follow the instructions. Therefore, the evaluation pipeline described here is necessary.}.
We then define accuracy as the percentage of outputs that the LLM judge considers aligned with the ground truth.
Please refer to Appendix~\ref{sec:appendix-llm-based-evaluation} for details on the prompts and the alignment between LLMs and humans.
\textbf{Importantly, although we used GPT-4o for evaluation in the main text, to support reproducibility we have also included comprehensive evaluation results using the open-source LLM (}\texttt{LLaMA3.1-8b-Instruct}\textbf{) in Appendix~\ref{sec:all-task-eval-results}.}

For regression tasks, the above method cannot be applied because performance is not assessed using hard labels.
Instead, we use LLMs (GPT-4o) as a \textit{post-processor} to transform natural language responses into a format compatible with the original metrics used in these tasks.
Specifically, we developed a prompt that directs the LLMs to extract essential information from the output and convert it into the same format as the ground truth.
If the output lacks correct or relevant information that can be converted to match the ground truth format, the LLM returns ``N/A'' to mark it as invalid.
Since conventional metrics cannot accommodate ``N/A'' and setting a default value is unreasonable due to the variability of metrics, we also calculate the N/A rate for each regression task, defined as the percentage of invalid outputs within a task.
A higher N/A rate indicates that the model struggles with following instructions, which may indirectly affect its performance on the task.

For sequence generation tasks such as ASR and captioning, we apply their original metrics directly to the raw outputs from baseline models.
This is because identifying redundant prefixes or sections unrelated to the task objectives is highly challenging in sequence generation, even with human involvement.
Besides, our review of the original evaluation results reported by these baseline models revealed no explicit mention of post-processing procedures.
Therefore, we base our evaluations solely on the unprocessed outputs, ensuring consistency and objectivity across all models.

\section{Results}

\begin{figure}[ht]
    \centering
    \includegraphics[width=.8\linewidth]{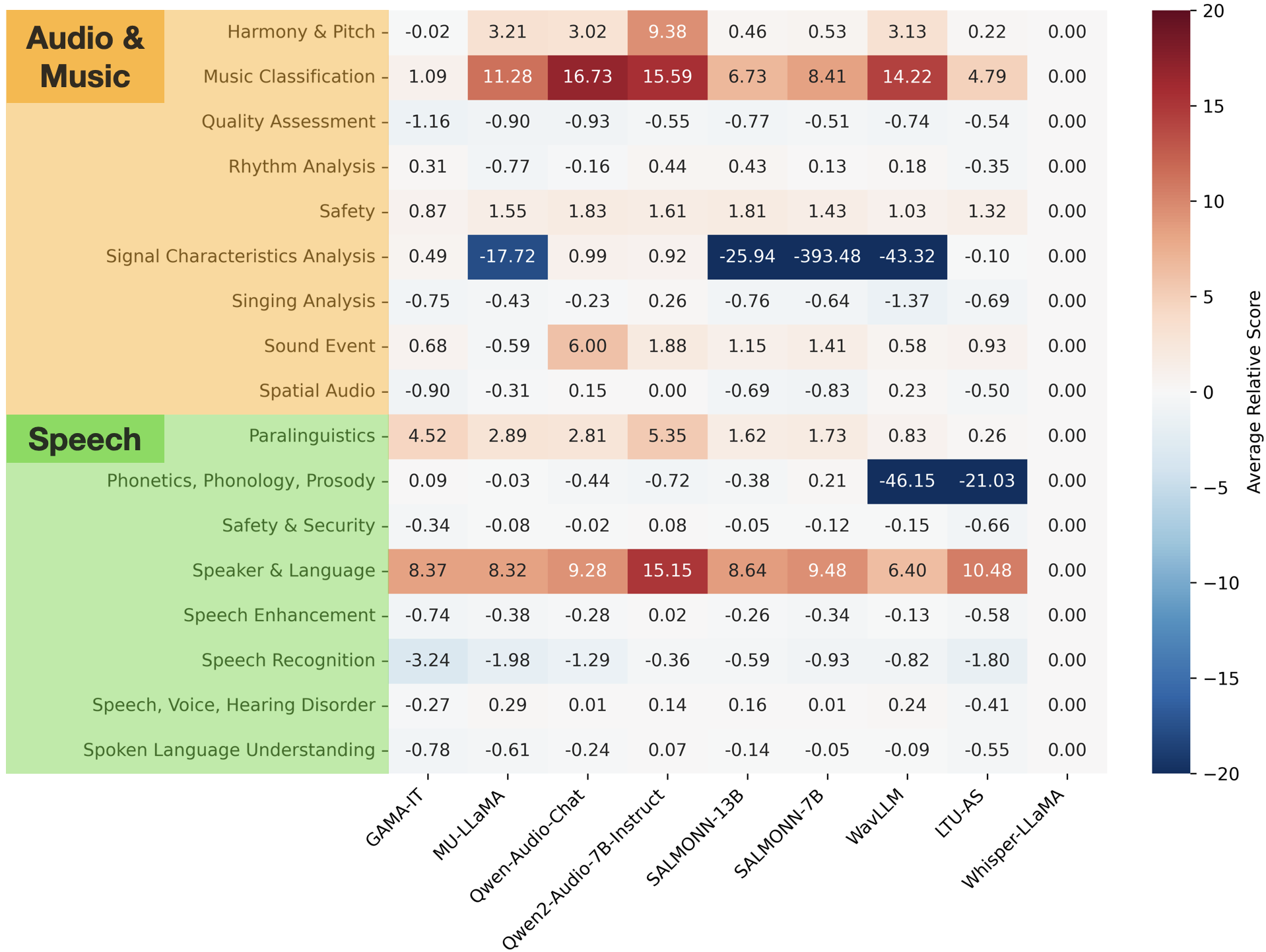}
    \caption{Domain-level relative-score-based performance comparison across different models. We exclude tasks where the models have a 100\% NA rate or where Whisper-LLaMA scores zero, as these make it impossible to average their scores with other tasks.}
    \label{fig:score-results}
\end{figure}

\subsection{Domain-Level Performance Comparison in Taxonomy}

Here, we analyze the results based on the taxonomy. Due to space limitations, we can only report and compare the results of different models at the domain level in this subsection.
Since different metrics are used across tasks, we adopted a \textbf{relative-score-based} method to summarize task performance.
For each task, we calculated the relative improvement of each model compared to the cascaded system baseline (Whisper + LLaMA) and then obtained the domain-level scores by averaging all its improvements across the tasks within each domain.
For regression tasks, we introduced the N/A rate to measure how well a model meets the task requirements.
The original task metric was only computed on instances that followed the task format (after post-processing by the LLM).
To account for whether a model can follow instructions, we incorporated the N/A rate into the reported scores.
Accordingly, we calculated the improvement based on scaled values.
For metrics where a higher value indicates better performance (such as the F1 score), we multiplied the metric value by (1 – N/A rate). Conversely, for metrics where a lower value indicates better performance (such as word error rate), we divided the metric value by (1 – N/A rate).

Figure~\ref{fig:score-results} presents a domain-level comparison of relative scores across different models.
Whisper-LLaMA consistently has zero scores throughout all domains, serving as the reference baseline for evaluating other models.
In speech domains, no model outperforms Whisper-LLaMA in speech recognition and spoken language understanding.
This shows that using ASR remains a strong baseline for language understanding, as text more explicitly represents semantic information than speech.
However, Whisper-LLaMA lags behind SALMONN in phoneme recognition, a task for which SALMONN is explicitly trained, indicating that task-specific training still provides superior performance in specialized domains.

Conversely, in the speaker and paralinguistics domains, all models surpass the baseline.
This improvement occurs because the ASR process in the cascaded system tends to discard critical information from the speech signal, such as speaker characteristics, pitch, and emotion.
In contrast, other models utilize soft representations that retain more speech information, giving them the potential to learn beyond ASR, and some of these models were also trained to perform specific tasks within certain domains
\footnote{Although the models have been trained on some tasks in the same domains, the training tasks of these models generally do not use the same instructions as Dynamic-SUPERB and do not offer as broad coverage.}.
Additionally, music-specific large language models like GAMA-IT and Mu-LLaMA perform poorly on speech recognition and understanding tasks, which is expected since they were primarily designed for music understanding.
Nevertheless, they surprisingly achieved scores comparable to the speech models in speaker and paralinguistics.

Turning to audio and music domains, where ASR models cannot transcribe non-speech information into text, most models outperform the baseline in several areas, such as music classification and sound event detection.
Surprisingly, spoken language models primarily trained on speech data (Qwen models, SALMONN, WavLLM)\footnote{Qwen and SALMONN are trained on a mixture of speech (dominant component), music, and audio data.} outperform the two music models (GAMA-IT and Mu-LLaMA) in various music-related domains such as harmony and pitch, music classification, and rhythm analysis.
We speculate that training on diverse data, even with significant differences in signal-level characteristics, enhances performance across domains, emphasizing the importance of developing unified models for speech, music, and general audio processing.

Lastly, we observe some outliers with significantly higher or lower values in the figure, typically resulting from unbounded evaluation metrics.
For instance, in the phonetics and prosody domain, WavLLM and LTU-AS exhibit poor scores due to their erroneous outputs in phone/phoneme segment counting tasks, predicting thousands of segments in utterances lasting only seconds and resulting in an excessively high mean square error.
While the relative score-based approach effectively summarizes model performance, domain-level scores can be distorted by specific tasks within a domain.
Thus, we encourage researchers aiming to develop model performance in specific domains to comprehensively report task-level scores to more accurately reflect their models' capabilities.

\vspace{-0.2em}

\subsection{Core tasks results}

\begin{table}[tb]
    \centering
    \caption{Evaluation on SUPERB tasks. A ``*'' indicates that the metric is scaled with NA rate, and a ``-'' means a model has a 100\% NA rate.}
    \label{tab:superb-results}
    \resizebox{\linewidth}{!}{
    \begin{tabular}{|c|r|r|r|r|r|r|r|r|r|r|}
         \hline
         {} & \multicolumn{1}{c|}{\textbf{PR}} & \multicolumn{1}{c|}{\textbf{KS}} & \multicolumn{1}{c|}{\textbf{IC}} & \multicolumn{1}{c|}{\textbf{ER}} & \multicolumn{1}{c|}{\textbf{ASR}} & \multicolumn{1}{c|}{\textbf{QbE}} & \multicolumn{2}{c|}{\textbf{SF}} & \multicolumn{1}{c|}{\textbf{SV}} & \multicolumn{1}{c|}{\textbf{SD}} \\
         \hline
         {} & PER $\downarrow$ & Acc $\uparrow$ & Acc $\uparrow$ & Acc $\uparrow$ & WER $\downarrow$ & Acc $\uparrow$ & F1$^*$ $\uparrow$ & CER$^*$ $\downarrow$ & Acc $\uparrow$ & DER$^*$ $\downarrow$ \\
         \hline
         \hline
         Whisper-LLaMA & 100.1 & 36.5 & 36.5 & 12.5 & 34.0 & 46.5 & 53.6 & 51.3 & 45.0 & 1068.7 \\
         SALMONN-7B & 25.4 & 30.5 & 21.0 & 12.1 & 15.1 & 49.0 & 61.6 & 61.5 & 45.0 & - \\
         SALMONN-13B & 24.6 & 2.0 & 5.5 & 12.5 & 2.8 & 51.5 & 29.6 & 106.3 & 51.0 & - \\
         Qwen-Audio-Chat & 100.6 & 60.5 & 26.5 & 70.4 & 69.4 & 48.0 & 42.2 & 90.1 & 49.0 & 8852.2 \\
         Qwen2-Audio-7B-Inst. & 101.0 & 47.0 & 26.0 & 75.8 & 36.7 & 53.5 & 42.4 & 55.7 & 51.0 & 4664.4 \\
         WavLLM & 100.0 & 43.0 & 28.0 & 12.1 & 6.9 & 49.5 & 50.8 & 84.9 & 49.0 & 3621.2 \\
         LTU-AS & 102.8 & 1.0 & 0.1 & 25.8 & 96.3 & 45.0 & 10.2 & 559.3 & 44.0 & 14923.3 \\
         GAMA-IT & 100.4 & 2.0 & 1.0 & 0.0 & 116.7 & 2.5 & 2.7 & 4750.0 & 45.0 & 187.6 \\
         Mu-LLaMA & 110.3 & 4.0 & 3.5 & 12.9 & 103.7 & 51.0 & 1.0 & 15869.6 & 44.0 & - \\
         \hline
    \end{tabular}
    }
    \justify
    {\footnotesize \textbf{PR}: Phoneme Recognition, \textbf{KS}: Keyword Spotting, \textbf{IC}: Intent Classification, \textbf{ER}: Emotion Recognition, \textbf{QbE}: Query-by-Example, \textbf{SF}: Slot Filling, \textbf{SV}: Speaker Verification, \textbf{SD}: Speaker Diarization}
    \vspace{-1.2em}
\end{table}

We tested the performance of the models on core tasks that are widely studied and commonly used.
Table~\ref{tab:superb-results} presents the detailed results of the models on SUPERB tasks, displaying each task's performance in its own metric rather than relative scores.
Please refer to Appendix~\ref{sec:appendix-hear-marble-results} for results on HEAR and MARBLE.
It is worth noting that these tasks have been adapted to the Dynamic-SUPERB framework.
Therefore, the results are not directly comparable to previous studies that tested on the original SUPERB.
However, they indicate current performance levels on these tasks.

No single model excels across all tasks.
In phoneme recognition (PR), the SALMONN models were the only ones to achieve a relatively lower phoneme error rate (PER) compared to the others.
For keyword spotting (KS), Qwen-Audio-Chat was the only model to perform slightly better than a random guess (50\%).
Whisper-LLaMA surpassed all other models in intent classification (IC).
Regarding emotion recognition (ER), Qwen-Audio-Chat and Qwen2-Audio-7B-Instruct stood out, with the latter even achieving about 75\% accuracy.
In ASR, SALMONN-13B and WavLLM are the only two models that achieved a word error rate (WER) lower than 10\%.
Query-by-example (QbE) appears to be very challenging for all models, as none performed better than a random guess (50\%).
Slot-filling (SF) is evaluated using two metrics: F1 for slot type and CER for slot value.
While Whisper-LLaMA and SALMONN-7B are the top two in this task, their results are still far from ideal.
In speaker verification (SV), all models performed badly, as even the best ones achieved results close to random guessing (50\%).
In speaker diarization (SD), results were similarly poor, with some even reaching a 100\% N/A rate.
Comparing Figure~\ref{fig:score-results} and Table~\ref{tab:superb-results}, we observed that performance on core tasks sometimes deviates from trends observed at the domain level.
For example, although SALMONN-13B outperforms other models in ASR, it lags behind Whisper-LLaMA in the speech recognition domain, revealing its limited capabilities and highlighting the need to enhance its generalizability.
In summary, among tested models, some excel in specific tasks, while others perform poorly across the board, and none dominate across all tasks.
We believe these findings provide valuable insights for researchers aiming to improve models, starting from core tasks and progressively tackling each domain in the benchmark for broader applicability.

\vspace{-0.4em}

\section{Conclusions}
This paper presents Dynamic-SUPERB \textit{Phase-2}, the largest benchmark for evaluating instruction-based universal spoken language models, building upon collaborative efforts across the research community.
Dynamic-SUPERB covers a wide range of diverse tasks and offers a fine-grained task taxonomy.
The recent models show good performance on specific tasks but poor generalization across common tasks like those in SUPERB, highlighting the need for further research on universal models.
All materials are made openly available to facilitate reproduction and benchmarking.
We sincerely invite researchers to join our vibrant community and collaborate to advance the field.

\textbf{Limitations}: 
Although Dynamic-SUPERB \textit{Phase-2} is the largest and most comprehensive benchmark, we acknowledge its limitations.
It lacks comprehensive speech-generation tasks, as  \textit{Phase-2} focused on understanding tasks due to the few universal generation models.
Despite our efforts to develop the task taxonomy scientifically, new domains may emerge as the benchmark grows, and tasks can be categorized in various ways.
While our automatic evaluation pipeline using LLMs correlates well with human evaluations (Appendix~\ref{sec:appendix-llm-based-evaluation}) for current tasks, it may not generalize to all future tasks.
Upholding the core spirit of the Dynamic-SUPERB project, we are striving to address these issues and enhance the benchmark for the next phase.

\section{Ethics Statement}
Dynamic-SUPERB \textit{Phase-2} includes several datasets, and we asked contributors to describe how they used them.
Most contributors derived task data by uniformly sampling from the original datasets without applying any special processing.
Therefore, Dynamic-SUPERB \textit{Phase-2} inevitably inherits the biases present in these datasets across tasks.
Additionally, we asked all contributors to provide the necessary license information to ensure that we can include them in the benchmark properly.
More specifically, all task data in Dynamic-SUPERB are derived from datasets with licenses that permit remixing and redistribution.
As we expect the benchmark to grow dynamically in the future, we maintain a complete list of dataset licenses on our official GitHub page (\url{https://github.com/dynamic-superb/dynamic-superb/blob/main/docs/dataset_license.md}), which will be updated as new tasks are introduced.
If there are any changes to the dataset license in the future, we may adjust the benchmark task accordingly.

\section{Reproducibility Statement}
We open-source all task data and the evaluation pipeline at \url{https://github.com/dynamic-superb/dynamic-superb}.
For LLM-based evaluation, we set GPT-4o's temperature to 0 to ensure a stable evaluation process.
However, we acknowledge that GPT-4o's behavior may change with future updates.
Therefore, we provide a comprehensive set of evaluation results using LLaMA-3.1-8b-Instruct in Appendix~\ref{sec:all-task-eval-results} (Table~\ref{tab:eval-result-llama}).
Additionally, in Appendix~\ref{sec:appendix-llm-based-evaluation}, we present a preliminary study on using open-source LLMs for evaluation, demonstrating promising correlations with human annotations.
We encourage researchers to use these open-source LLMs when evaluating their own models to support reproducibility.

\section{Acknowledgement}
We thank the National Center for High-performance Computing (NCHC) of National Applied Research Laboratories (NARLabs) in Taiwan for providing computational and storage resources.

\bibliography{iclr2025_conference}
\bibliographystyle{iclr2025_conference}

\appendix
\newpage
\section{Task taxonomy}
\label{sec:appendix-task-taxonomy}

\begin{figure}[H]
    \centering
    \includegraphics[width=\linewidth]{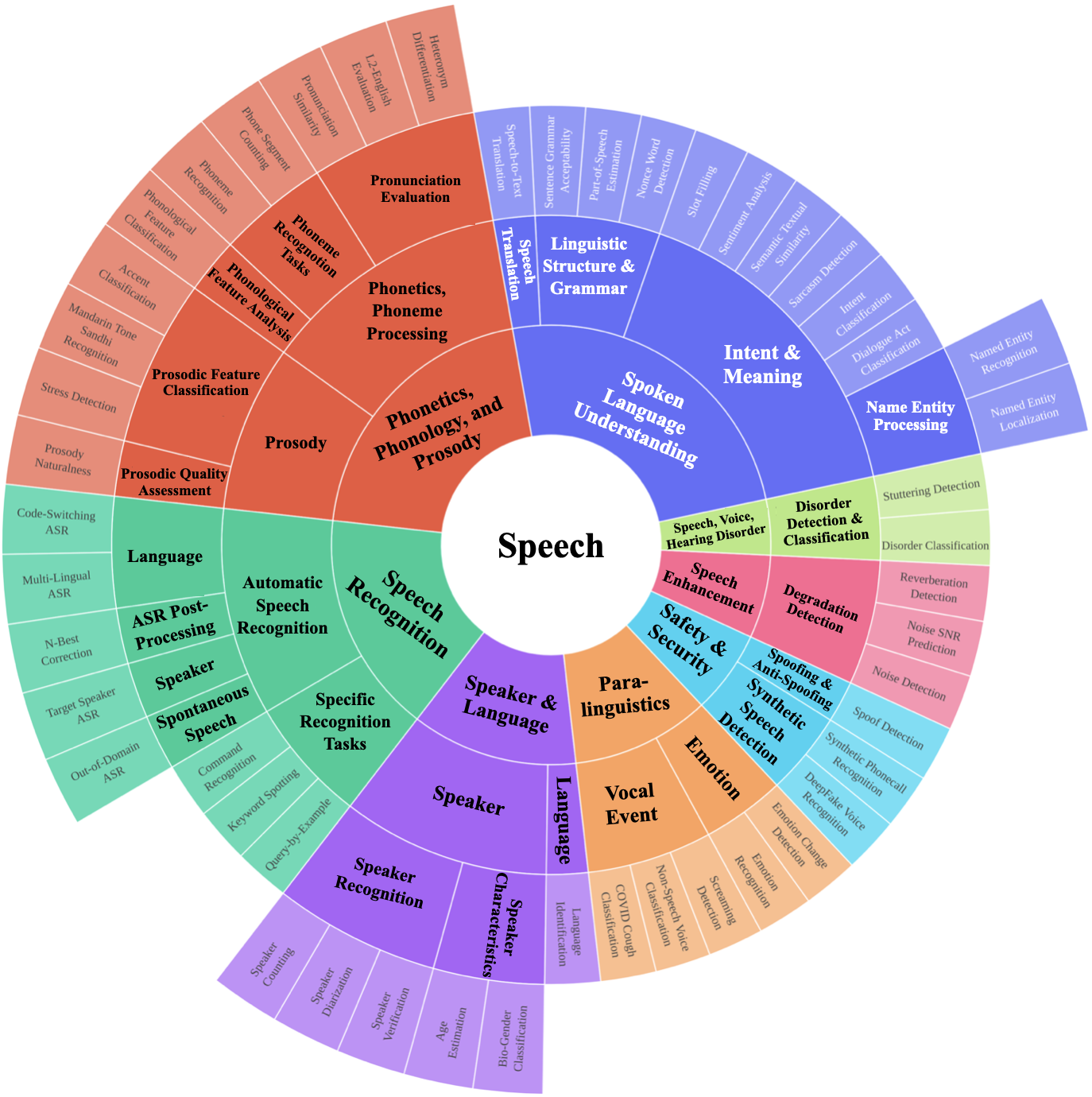}
    \caption{Task taxonomy of speech.}
    \label{fig:complete-speech-taxonomy}
\end{figure}

Figures \ref{fig:complete-speech-taxonomy} and \ref{fig:complete-audio-taxonomy} show the complete task taxonomy in Dynamic-SUPERB \textit{Phase-2}.
Additionally, Table~\ref{tab:appendix-taxonomy-comparison} compares our task taxonomy with the INTERSPEECH conference and EDICS from IEEE SPS.
While not identical, our taxonomy is well-aligned with those organized by professional conferences and institutions.
\vfill
\vfill

\newpage
\begin{center}             
    \vspace*{\fill}        
    \begin{figure}[H]
        \centering
        \includegraphics[width=\linewidth]{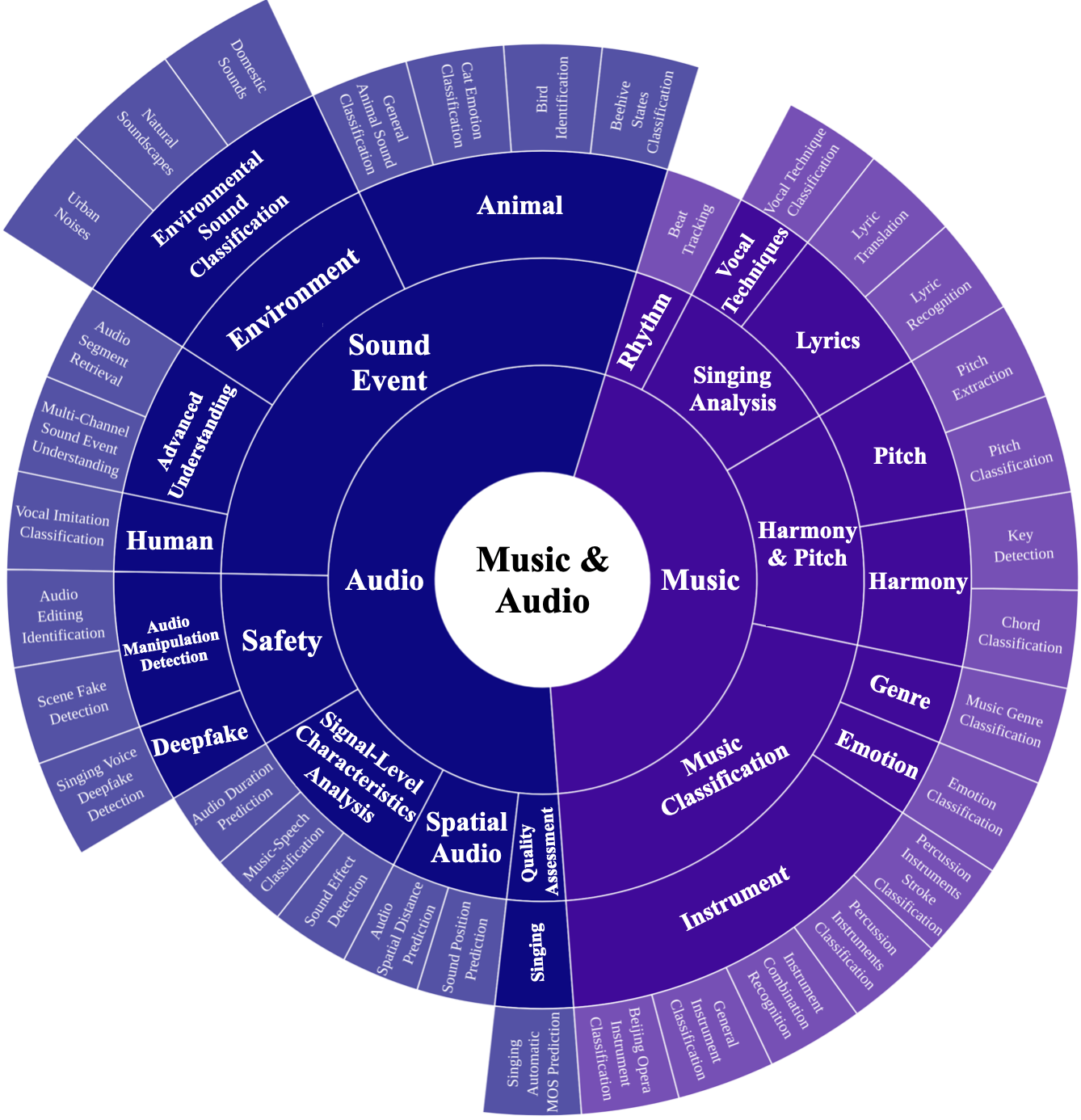}
        \caption{Task taxonomy of audio and music.}
        \label{fig:complete-audio-taxonomy}
    \end{figure}
    \vspace*{\fill}        
\end{center}

\newpage
\begin{table}[H]
    \centering
    \caption{Comparison of task taxonomy in Dynamic-SUPERB \textit{Phase-2} with INTERSPEECH and EDICS. A ``\checkmark'' indicates that a conference session, area, or EDICS category matches or closely aligns with the taxonomy.}
    \label{tab:appendix-taxonomy-comparison}
\begin{tabular}{lcc}
         \toprule
         \textbf{Taxonomy} & \textbf{INTERSPEECH} & \textbf{EDICS} \\
         \midrule
         Music/Harmony \& Pitch & {} & \checkmark \\
         Music/Music Classification & {} & {} \\
         Music/Rhythm Analysis & {} & \checkmark \\
         Audio/Quality Assessment & {} & \checkmark \\
         Audio/Safety & {} & \checkmark \\
         Audio/Signal-Characteristics Analysis & {} & {} \\
         Audio/Singing Analysis & \checkmark & {} \\
         Audio/Sound Event & \checkmark & \checkmark \\
         Audio/Spatial Audio Analysis & \checkmark & \checkmark \\
         Speech/Paralinguistics & \checkmark & {} \\
         Speech/Phonetics, Phonology, Prosody & \checkmark & {} \\
         Speech/Safety and Security & \checkmark & {} \\
         Speech/Speaker \& Language & \checkmark & {} \\
         Speech/Speech Enhancement & \checkmark & \checkmark \\
         Speech/Speech Recognition & \checkmark & \checkmark \\
         Speech/Speech, Voice, Hearing Disorder & \checkmark & {} \\
         Speech/Spoken Language Understanding & \checkmark & \checkmark \\
         \bottomrule
         
    \end{tabular}
\end{table}

\section{Dynamic-SUPERB tasks}
\label{sec:complete-task-list}

\begin{center}
\begin{longtable}{
>{\raggedright\arraybackslash}p{3.5cm}|>{\raggedright\arraybackslash}p{3.0cm}|>{\raggedright\arraybackslash}p{5.5cm}
}
    \caption{The list of all tasks in Dynamic-SUPERB \textit{Phase-2}. Tasks are ordered by the taxonomy. Core tasks are highlighted in green (SUPERB), blue (HEAR), or orange (MARBLE).} \label{tab:speech-task-list} \\
    \toprule
    \textbf{Domain} & \textbf{Task} & \textbf{Dataset} \\
    \midrule
    \endfirsthead
    
    \toprule
    \textbf{Domain} & \textbf{Task} & \textbf{Dataset} \\
    \midrule
    \endhead
    
    \midrule
    \multicolumn{3}{r}{\textit{Continued on next page}} \\
    \endfoot
    
    \bottomrule
    \endlastfoot

    Audio / Harmony \& Pitch / Harmony & Chord Classification  & Musical Instrument Chord Classification (Audio) \citep{kaggle2024musicalinstruments} \\
    \midrule
    \rowcolor{YellowOrange!60}
    Audio / Harmony \& Pitch / Harmony & MARBLE Key Detection & GiantSteps (key) \citep{knees2015two} \\
    \midrule
    \rowcolor{SkyBlue!60}
    Audio / Harmony \& Pitch / Pitch & HEAR Music Transcription & MAESTRO \citep{hawthorne2018enabling} \\
    \midrule
    \rowcolor{SkyBlue!60}
    Audio / Harmony \& Pitch / Pitch & HEAR Percussion Instruments Tonic Classification & Mridangam Stroke \citep{anantapadmanabhan2013modal} \\
    \midrule
    Audio / Harmony \& Pitch / Pitch & Instrument Pitch Classification & Nsynth \citep{engel2017neural}  \\
    \midrule
    Audio / Harmony \& Pitch / Pitch & Pitch Extraction By Lyrics & CSD \citep{choi2020children} \\
    \midrule
    Audio / Music Classification / Emotion & Emotion Classificaiton In Songs & EMOTIFY \citep{ALJANAKI2016115} \\
    \midrule
    \rowcolor{YellowOrange!60}
    Audio / Music Classification / Emotion & MARBLE Emotion Detection & MTG-Jamendo \citep{bogdanov2019mtg} \\
\bottomrule
    
\end{longtable}
\end{center}

\section{Task-level evaluation results}
\label{sec:all-task-eval-results}
\begin{center}
\begin{longtable}{
>{\raggedright\arraybackslash}p{1.6cm}|
>{\raggedright\arraybackslash}p{0.8cm}|
*{9}{>{\raggedright\arraybackslash}p{0.7cm}}
}
    \caption{Task-level Evaluation Results: GPT4o} 
    \label{tab:eval-result-gpt4o} \\
    \toprule
        {\tiny \textbf{Task}} & {\tiny \textbf{Metric}} & 
        \fontsize{5pt}{5pt}\selectfont \textbf{Whisper-LLaMA} & {\fontsize{5pt}{5pt}\selectfont \textbf{LTU-AS}} & 
        \fontsize{5pt}{5pt}\selectfont \textbf{SALMONN 7B} & \fontsize{5pt}{5pt}\selectfont \textbf{SALMONN 13B} & 
        \fontsize{5pt}{5pt}\selectfont \textbf{Qwen-Audio-Chat} & \fontsize{5pt}{5pt}\selectfont \textbf{Qwen2-Audio-7B-Inst.} & 
        {\fontsize{5pt}{5pt}\selectfont \textbf{WavLLM}} & 
        \fontsize{5pt}{5pt}\selectfont \textbf{MU-LLaMA} & \fontsize{5pt}{5pt}\selectfont \textbf{GAMA-IT} \\

    \midrule
    \endfirsthead
    
    \toprule
        {\tiny \textbf{Task}} & {\tiny \textbf{Metric}} & 
        \fontsize{5pt}{5pt}\selectfont \textbf{Whisper-LLaMA} & {\fontsize{5pt}{5pt}\selectfont \textbf{LTU-AS}} & 
        \fontsize{5pt}{5pt}\selectfont \textbf{SALMONN 7B} & \fontsize{5pt}{5pt}\selectfont \textbf{SALMONN 13B} & 
        \fontsize{5pt}{5pt}\selectfont \textbf{Qwen-Audio-Chat} & \fontsize{5pt}{5pt}\selectfont \textbf{Qwen2-Audio-7B-Inst.} & 
        {\fontsize{5pt}{5pt}\selectfont \textbf{WavLLM}} & 
        \fontsize{5pt}{5pt}\selectfont \textbf{MU-LLaMA} & \fontsize{5pt}{5pt}\selectfont \textbf{GAMA-IT} \\
    \midrule
    \endhead
    
    \midrule
    \multicolumn{11}{r}{\textit{Continued on next page}} \\
    \endfoot
    
    \bottomrule
    \endlastfoot
    
    \tiny Chord Classification & \tiny LLM-C$\uparrow$ & \tiny 4.00\% & \tiny 10.00\% & \tiny 0.50\% & \tiny 9.00\% & \tiny 44.00\% & \tiny 28.50\% & \tiny 44.50\% & \tiny 51.00\% & \tiny 0.00\%\\
    \midrule
    \tiny MARBLE Key Detection (GiantSteps - Key) & \tiny LLM-C$\uparrow$ & \tiny 3.00\% & \tiny 0.50\% & \tiny 1.50\% & \tiny 1.00\% & \tiny 3.00\% & \tiny 17.00\% & \tiny 2.00\% & \tiny 0.50\% & \tiny 0.00\%\\
    \midrule
    \tiny HEAR Music Transcription (MAESTRO) & \tiny NAR$\downarrow$ & \tiny 92.43\% & \tiny 100.00\% & \tiny 99.46\% & \tiny 100.00\% & \tiny 96.76\% & \tiny 96.76\% & \tiny 96.76\% & \tiny 100.00\% & \tiny 100.00\%\\
    \midrule
    \tiny HEAR Music Transcription (MAESTRO) & \tiny TER$\downarrow$ & \tiny 1.0065 & \tiny - & \tiny 36.0000 & \tiny - & \tiny 8.3438 & \tiny 1.6875 & \tiny 6.25 & \tiny - & \tiny -\\
    \midrule
    \tiny HEAR Percussion Instruments Tonic Classification (Mridangam Tonic) & \tiny LLM-C$\uparrow$ & \tiny 2.00\% & \tiny 2.50\% & \tiny 12.70\% & \tiny 13.40\% & \tiny 11.30\% & \tiny 11.80\% & \tiny 12.80\% & \tiny 12.50\% & \tiny 5.70\%\\
    \midrule
    \tiny Instrument Pitch Classification (Nsynth) & \tiny LLM-C$\uparrow$ & \tiny 0.56\% & \tiny 0.56\% & \tiny 0.11\% & \tiny 0.22\% & \tiny 0.67\% & \tiny 18.11\% & \tiny 0.78\% & \tiny 0.33\% & \tiny 0.78\%\\
    \midrule
    \tiny Pitch Extraction By Lyrics (CSD) & \tiny NAR$\downarrow$ & \tiny 37.50\% & \tiny 94.00\% & \tiny 47.50\% & \tiny 73.50\% & \tiny 65.50\% & \tiny 84.50\% & \tiny 76.00\% & \tiny 98.00\% & \tiny 68.50\%\\
    \midrule
    \tiny Pitch Extraction By Lyrics (CSD) & \tiny TER$\downarrow$ & \tiny 146.28\% & \tiny 116.67\% & \tiny 226.21\% & \tiny 639.24\% & \tiny 106.45\% & \tiny 165.96\% & \tiny 136.11\% & \tiny 100.00\% & \tiny 195.18\%\\
\bottomrule
\end{longtable}
\end{center}
\begin{center}
    \begin{longtable}{
    >{\raggedright\arraybackslash}p{1.6cm}|
    >{\raggedright\arraybackslash}p{0.8cm}|
    *{9}{>{\raggedright\arraybackslash}p{0.7cm}}
    }
        \caption{Task-level evaluation results using LLaMA-3.1-8B-Instruct.} 
        \label{tab:eval-result-llama} \\
        \toprule
        {\tiny \textbf{Task}} & {\tiny \textbf{Metric}} & 
        \fontsize{5pt}{5pt}\selectfont \textbf{Whisper-LLaMA} & {\fontsize{5pt}{5pt}\selectfont \textbf{LTU-AS}} & 
        \fontsize{5pt}{5pt}\selectfont \textbf{SALMONN 7B} & \fontsize{5pt}{5pt}\selectfont \textbf{SALMONN 13B} & 
        \fontsize{5pt}{5pt}\selectfont \textbf{Qwen-Audio-Chat} & \fontsize{5pt}{5pt}\selectfont \textbf{Qwen2-Audio-7B-Inst.} & 
        {\fontsize{5pt}{5pt}\selectfont \textbf{WavLLM}} & 
        \fontsize{5pt}{5pt}\selectfont \textbf{MU-LLaMA} & \fontsize{5pt}{5pt}\selectfont \textbf{GAMA-IT} \\

        \midrule
        \endfirsthead
        
        \toprule
        {\tiny \textbf{Task}} & {\tiny \textbf{Metric}} & 
        \fontsize{5pt}{5pt}\selectfont \textbf{Whisper-LLaMA} & {\fontsize{5pt}{5pt}\selectfont \textbf{LTU-AS}} & 
        \fontsize{5pt}{5pt}\selectfont \textbf{SALMONN 7B} & \fontsize{5pt}{5pt}\selectfont \textbf{SALMONN 13B} & 
        \fontsize{5pt}{5pt}\selectfont \textbf{Qwen-Audio-Chat} & \fontsize{5pt}{5pt}\selectfont \textbf{Qwen2-Audio-7B-Inst.} & 
        {\fontsize{5pt}{5pt}\selectfont \textbf{WavLLM}} & 
        \fontsize{5pt}{5pt}\selectfont \textbf{MU-LLaMA} & \fontsize{5pt}{5pt}\selectfont \textbf{GAMA-IT} \\
        \midrule
        \endhead
        
        \midrule
        \multicolumn{11}{r}{\textit{Continued on next page}} \\
        \endfoot
        
        \bottomrule
        \endlastfoot
        
            \tiny Chord Classification & \tiny LLM-C$\uparrow$ & \tiny 4.00\% & \tiny 5.00\% & \tiny 7.50\% & \tiny 13.00\% & \tiny 43.00\% & \tiny 16.00\% & \tiny 45.50\% & \tiny 50.00\% & \tiny 0.50\%\\
        \midrule
    
        \tiny MARBLE Key Detection (GiantSteps - Key) & \tiny LLM-C$\uparrow$ & \tiny 3.50\% & \tiny 1.00\% & \tiny 1.50\% & \tiny 1.00\% & \tiny 3.00\% & \tiny 13.00\% & \tiny 2.00\% & \tiny 0.50\% & \tiny 11.50\%\\
        \midrule
    
        \tiny HEAR Music Transcription (MAESTRO) & \tiny NAR$\downarrow$ & \tiny 95.14\% & \tiny 100.00\% & \tiny 100.00\% & \tiny 100.00\% & \tiny 97.30\% & \tiny 96.22\% & \tiny 84.86\% & \tiny 100.00\% & \tiny 100.00\%\\
        \midrule
    
        \tiny HEAR Music Transcription (MAESTRO) & \tiny TER$\downarrow$ & \tiny 99.45\% & \tiny - & \tiny - & \tiny - & \tiny 99.54\% & \tiny 100.00\% & \tiny 99.48\% & \tiny - & \tiny -\\
        \midrule
    
        \tiny HEAR Percussion Instruments Tonic Classification (Mridangam Tonic) & \tiny LLM-C$\uparrow$ & \tiny 2.40\% & \tiny 3.30\% & \tiny 14.00\% & \tiny 15.10\% & \tiny 14.00\% & \tiny 13.20\% & \tiny 16.00\% & \tiny 14.30\% & \tiny 8.60\%\\
        \midrule
    
        \tiny Instrument Pitch Classification (Nsynth) & \tiny LLM-C$\uparrow$ & \tiny 0.70\% & \tiny 0.90\% & \tiny 1.70\% & \tiny 5.40\% & \tiny 1.60\% & \tiny 18.90\% & \tiny 1.20\% & \tiny 0.40\% & \tiny 1.00\%\\
        \midrule
    
        \tiny Pitch Extraction By Lyrics (CSD) & \tiny NAR$\downarrow$ & \tiny 30.00\% & \tiny 73.50\% & \tiny 57.00\% & \tiny 85.50\% & \tiny 76.50\% & \tiny 73.00\% & \tiny 57.50\% & \tiny 97.50\% & \tiny 53.00\%\\
        \midrule
    
        \tiny Pitch Extraction By Lyrics (CSD) & \tiny TER$\downarrow$ & \tiny 1.86 & \tiny 2.82 & \tiny 4.32 & \tiny 7.38 & \tiny 2.33 & \tiny 2.42 & \tiny 2.71 & \tiny 1.00 & \tiny 3.87\\
\bottomrule
    \end{longtable}
    \end{center}

\vspace{-3em}
\section{HEAR \& MARBLE Evaluation Results}
\label{sec:appendix-hear-marble-results}

Tables~\ref{tab:hear_table} and \ref{tab:marble_table} show the evaluation results for the HEAR and MARBLE tasks, respectively, as collected in Dynamic-SUPERB \textit{Phase-2}.
Importantly, we replaced the datasets for some tasks to address licensing issues and reduced their size to enable more efficient inference.
Furthermore, we removed certain tasks because they overlap with existing SUPERB tasks.
Consequently, these results are not directly comparable to those evaluated on the original HEAR and MARBLE.

\begin{table}[H]
    \centering
    \caption{HEAR (audio) Results. \textit{LLM-C} stands for LLM Classification, \textit{NAR} stands for Not Applicable Rate, \textit{TER} stands for Token Error Rate, and \textit{ACC} stands for Accuracy.}
    \label{tab:hear_table}
    \resizebox{\linewidth}{!}{
    \begin{tabular}{c|c|cccccccccc}
         \toprule
         \multirow{2}{*}{\textbf{Task}} & \multirow{2}{*}{\textbf{Metric}} & \textbf{Whisper-} & \multirow{2}{*}{\textbf{LTU-AS}} & \textbf{SALMONN} & \textbf{SALMONN} & \textbf{Qwen-} & \textbf{Qwen2-} & \multirow{2}{*}{\textbf{WavLLM}} & \textbf{MU-} & \multirow{2}{*}{\textbf{GAMA-IT}} \\
         {} & {} & {\textbf{LLaMA}} & {} & \textbf{7B} & \textbf{13B} & \textbf{Audio-Chat} & \textbf{Audio-7B-Inst.} & {} & \textbf{LLaMA} & {} \\
         \midrule
         \midrule
         \makecell{Beehive States Classification \\ - Beehive States} & LLM-C $\uparrow$ & 43.00\% & 35.00\% & 54.00\% & 42.50\% & 19.00\% & 42.50\% & 37.00\% & 37.50\% & 19.00\% \\
         \midrule
         \makecell{Emotion Recognition \\ - CREMAD} & LLM-C $\uparrow$ & 6.70\% & 9.30\% & 18.80\% & 18.10\% & 62.50\% & 61.10\% & 25.60\% & 12.40\% & 13.20\% \\
         \midrule
         \makecell{Environmental Sound \\ Classification - ESC-50} & LLM-C $\uparrow$ & 4.80\% & 25.30\% & 61.00\% & 73.40\% & 68.80\% & 43.90\% & 9.80\% & 0.40\% & 35.30\% \\
         \midrule
         \makecell{Language Identification \\ - VoxLingua107 Top10} & LLM-C $\uparrow$ & 92.18\% & 1.95\% & 6.28\% & 11.01\% & 18.00\% & 36.11\% & 1.85\% & 0.00\% & 0.10\% \\
         \midrule
         \makecell{Music Genre Classification \\ - ISMIR04} & LLM-C $\uparrow$ & 15.58\% & 7.04\% & 31.16\% & 44.22\% & 33.67\% & 37.69\% & 16.58\% & 27.64\% & 2.51\% \\
         \midrule
         \makecell{Music Speech Classification \\ - MAESTRO-LibriSpeech} & LLM-C $\uparrow$ & 72.50\% & 77.50\% & 91.50\% & 99.50\% & 98.00\% & 89.50\% & 59.50\% & 51.50\% & 88.00\% \\
         \midrule
         \multirow{2}{*}{\makecell{Music Transcription \\ - MAESTRO}} & NAR $\downarrow$ & 92.43\% & 100.00\% & 99.46\% & 100.00\% & 96.76\% & 96.76\% & 96.76\% & 100.00\% & 100.00\%  \\
         {} & TER $\downarrow$ & 1.0065 & - & 36.0000 & - & 8.3438 & 1.6875 & 6.2500 & - & - \\
         \midrule
         \makecell{Percussion Instruments \\ Classification - Beijing Opera \\ Percussion Instrument} & LLM-C $\uparrow$ & 3.39\% & 4.24\% & 10.17\% & 20.76\% & 16.10\% & 27.12\% & 24.15\% & 22.03\% & 11.44\% \\
         \midrule
         \makecell{Percussion Instruments \\ Stroke Classification \\ - Mridangam Stroke} & LLM-C $\uparrow$ & 0.10\% & 3.80\% & 7.60\% & 4.90\% & 11.50\% & 10.80\% & 13.70\% & 10.30\% & 0.10\% \\
         \midrule
         \makecell{Percussion Instruments \\ Tonic Classification \\ - Mridangam Stroke} & LLM-C $\uparrow$ & 2.00\% & 2.50\% & 12.70\% & 13.40\% & 11.30\% & 11.80\% & 12.80\% & 12.50\% & 5.70\% \\
         \midrule
         \makecell{Sound Event Detection \\ - DCASE2016 Task2} & LLM-C $\uparrow$ & 0.00\% & 0.00\% & 21.43\% & 35.71\% & 7.14\% & 14.29\% & 21.43\% & 21.43\% & 7.14\% \\
         \midrule
         \multirow{2}{*}{\makecell{Speaker Count Identification \\ - LibriCount}} & NAR $\downarrow$ & 30.40\% & 42.10\% & 35.90\% & 42.50\% & 3.50\% & 0.40\% & 22.00\% & 23.20\% & 26.20\% \\
         {} & ACC $\uparrow$ & 17.53\% & 16.93\% & 13.26\% & 17.22\% & 10.26\% & 13.76\% & 5.51\% & 13.80\% & 17.48\% \\
         \midrule
         \makecell{Vocal Imitation Classification \\ - Vocal Imitations} & LLM-C $\uparrow$ & 15.17\% & 5.50\% & 1.17\% & 1.83\% & 18.67\% & 11.17\% & 17.50\% & 3.83\% & 5.83\% \\
         \bottomrule
    \end{tabular}
    }
\end{table}

\begin{table}[H]
    \centering
    \caption{MARBLE (music) Results. \textit{LLM-C} stands for LLM Classification, \textit{NAR} stands for Not Applicable Rate, and \textit{MSE} stands for Mean Squared Error.}
    \label{tab:marble_table}
    \resizebox{\linewidth}{!}{
    \begin{tabular}{c|c|cccccccccc}
         \toprule
         \multirow{2}{*}{\textbf{Task}} & \multirow{2}{*}{\textbf{Metric}} & \textbf{Whisper-} & \multirow{2}{*}{\textbf{LTU-AS}} & \textbf{SALMONN} & \textbf{SALMONN} & \textbf{Qwen-} & \textbf{Qwen2-} & \multirow{2}{*}{\textbf{WavLLM}} & \textbf{MU-} & \multirow{2}{*}{\textbf{GAMA-IT}} \\
         {} & {} & {\textbf{LLaMA}} & {} & \textbf{7B} & \textbf{13B} & \textbf{Audio-Chat} & \textbf{Audio-7B-Inst.} & {} & \textbf{LLaMA} & {} \\
         \midrule
         \makecell{Key Detection \\ - Giantsteps (Key)} & LLM-C $\uparrow$ & 3.00\% & 0.50\% & 1.50\% & 1.00\% & 3.00\% & 17.00\% & 2.00\% & 0.50\% & 0.00\% \\
         \midrule
         \makecell{Music Tagging \\ - MTG-Jamendo} & LLM-C $\uparrow$ & 0.10\% & 0.30\% & 0.00\% & 0.00\% & 2.90\% & 0.30\% & 0.10\% & 0.50\% & 0.10\% \\
         \midrule
         \makecell{Music Tagging \\ - MagnaTagATune} & LLM-C $\uparrow$ & 5.50\% & 1.00\% & 3.50\% & 9.00\% & 4.50\% & 1.50\% & 1.00\% & 3.50\% & 7.50\% \\
         \midrule
         \makecell{Genre Classification \\ - MTG-Jamendo} & LLM-C $\uparrow$ & 0.00\% & 1.60\% & 0.10\% & 0.20\% & 3.20\% & 0.10\% & 0.20\% & 0.20\% & 0.00\% \\
         \midrule
         \makecell{Emotion Detection \\ - MTG-Jamendo} & LLM-C $\uparrow$ & 0.00\% & 1.50\% & 0.00\% & 0.00\% & 2.70\% & 0.80\% & 0.80\% & 0.80\% & 0.10\% \\
         \midrule
         \multirow{2}{*}{\makecell{Beat Tracking \\ - ASAP}} & NAR $\downarrow$ & 100.00\% & 100.00\% & 100.00\% & 100.00\% & 100.00\% & 100.00\% & 100.00\% & 100.00\% & 100.00\% \\
         {} & MSE $\uparrow$ & - & - & - & - & - & - & - & - & -\\
         \midrule
         \makecell{Vocal Technique \\ Detection - VocalSet} & LLM-C $\uparrow$ & 8.00\% & 1.00\% & 11.00\% & 6.50\% & 6.50\% & 15.50\% & 8.00\% & 8.50\% & 2.50\% \\
         \midrule
         \makecell{Instrument Classification \\ - MTG-Jamendo} & LLM-C $\uparrow$ & 0.40\% & 2.30\% & 0.00\% & 0.00\% & 6.40\% & 1.40\% & 0.20\% & 0.60\% & 0.20\% \\
         \bottomrule
    \end{tabular}
    }
\end{table}

\newpage
\section{LLM-based evaluation}
\label{sec:appendix-llm-based-evaluation}

Here, we list the prompt templates used in the evaluation. Figure~\ref{fig:classification-prompt} illustrates the prompt for classification task evaluation, while Figure~\ref{fig:regression-prompt} shows the prompt for regression task evaluation.
In addition to GPT-4o, we also employed open-source LLMs to ensure better reproducibility. Tables~\ref{tab:classification_results} and ~\ref{tab:regression_results} present the performance evaluations of the proposed prompts on classification and regression tasks, respectively, comparing their correlation with a human-labeled evaluation set. We found that GPT-4o and LLaMA-3.1-70B-Instruct achieved very high accuracy in both classification and regression tasks. This demonstrates that our evaluation using GPT-4o is reliable and that researchers can leverage LLaMA-3.1-70B-Instruct to conduct fully reproducible evaluations.

\begin{table}[H]
\centering
\caption{Performance evaluation of the proposed post-processing prompt on \textbf{classification} tasks using a human-labeled evaluation set. Accuracy is determined by agreement between the model output and the corresponding human label.}
\label{tab:classification_results}
\resizebox{.675\textwidth}{!}{
\begin{tabular}{l|c|c}
\toprule
\textbf{Model Name} & \textbf{Accuracy} & \textbf{Correct Count} \\ \midrule
gpt-4o & \textbf{98.67\%} & \textbf{148/150} \\
gpt-4o-mini & 98.00\% & 147/150 \\ \midrule
gpt-o1-preview & 98.67\% & 148/150 \\
gpt-o1-mini & \textbf{99.33\%} & \textbf{149/150} \\ \midrule
llama-3.1-70B-Instruct & \textbf{96.67\%} & \textbf{145/150} \\
llama-3.1-8B-Instruct & 92.67\% & 139/150 \\ \midrule
llama-3.2-3B-Instruct & 86.67 \% & 130/150 \\
llama-3.2-1B-Instruct & 61.33\% & 92/150 \\ \bottomrule
\end{tabular}
}
\end{table}

\begin{table}[H]
\centering
\caption{Performance evaluation of the proposed post-processing prompt on \textbf{regression} tasks using a human-labeled evaluation set. Accuracy is determined by exact matches between the post-processed model output and the corresponding human-generated post-processed output.}
\resizebox{.675\textwidth}{!}{
\begin{tabular}{l|c|c}
\toprule
\textbf{Model Name} & \textbf{Accuracy} & \textbf{Correct Count} \\ \midrule
gpt-4o & \textbf{92.00\%} & \textbf{138/150} \\
gpt-4o-mini & 86.67\% & 130/150 \\ \midrule
gpt-o1-preview & 86.00\% & 129/150 \\
gpt-o1-mini & 89.33\% & 134/150 \\ \midrule
llama-3.1-70B-Instruct & \textbf{90.00\%} & \textbf{135/150} \\
llama-3.1-8B-Instruct & 82.00\% & 123/150 \\ \midrule
llama-3.2-3B-Instruct & 61.33\% & 92/150 \\
llama-3.2-1B-Instruct & 39.33\% & 59/150 \\ \bottomrule
\end{tabular}
}
\label{tab:regression_results}
\end{table}

\begin{figure}[H]
    \centering
    \fbox{\parbox{0.95\columnwidth}{
        \small
        \textbf{Classification Post-processing Prompt:}\\
        You will be given a question, a corresponding correct answer(s), and a response from a model.
        The model's response is a reply to the question. Your task is to judge if the "Model's Response" aligns with the "Ground Truth Answer" based on the "Question." \\
        
        Please strictly follow the guidelines below:\\
        - Briefly explain the reasons for your judgment.\\
        - Answer with the format "Result: $<$YES or NO$>$" at the end.\\
        - Output "YES" if the response aligns with the ground truth answer; output "NO" if the response does not match the ground truth answer, selects incorrect or irrelevant options, or provides more answers than required.\\
        - The questions would be single-choice or multi-choice:\\
        For single-choice questions, the model's response should contain one and only one answer. If the model's response selects more than one answer or does not clearly indicate a single answer, you should mark it as incorrect and output "NO."
        For multi-choice questions, the model's response must exactly match all applicable correct choices. If the model's response selects too many, too few, or any incorrect answers, you should mark it as incorrect and output "NO."\\
        - Since the question is short answer, the model's response does not need to mention the content of the question. You only need to check if the model's response has the same meaning as the ground truth answer(s).\\

        Input Format:\\
        Question: \{instruction\}\\
        Ground Truth Answer: \{label\}\\
        Model's Response: \{response\}
    }}
    \caption{The prompt used for post-processing the \textbf{classification} tasks inference outputs in the study.}
    \label{fig:classification-prompt}
\end{figure}

\begin{figure}[H]
    \centering
    \fbox{\parbox{0.95\columnwidth}{
        \small
        \textbf{Regression Post-processing Prompt:}\\
        You will be provided with an "instruction," a "ground-truth label," and a "model output."
        Your job is to extract the value from the given "model output" and return the post-processed "model output".
        Provide your output with just the post-processed output without an explanation.\\

        [Task Description]\\
        - Analyze the "instruction" and "ground-truth label" to understand the desired output format.\\
        - Post-process the "model output" so that it matches the format of the "ground-truth label" and is numerically comparable to it.\\

        [Guidelines]\\
        - If the "model output" can be adjusted to match the format of the "ground-truth label," return only the post-processed result.\\
        - If the "model output" is already in an appropriate format, return it as is without changing it.\\
        - If it can not be adjusted into the correct format, does not clearly indicate any of the candidates, is ambiguous, or implies more than one answer, return "N/A."\\
        - Do not confuse the "ground-truth label" with "model output"; you are only supposed to use the "ground-truth label" to understand the desired output format.\\
        - Do not tamper with the original content of "model output"; you are only supposed to post-process the format of "model output" so it matches the format of "ground-truth label" and can be compared or evaluated against each other directly.\\
        - The post-processed output should only be a number in numeric form.\\
    
        Below you have three examples. Please consider their patterns to understand your task better:\\
        
        [Example 1 - Inputs]\\
        - "instruction": "Please identify the total times of code-switching in this wavefile."\\
        - "ground-truth label": 8\\
        - "model output": "There are no instances of code-switching in this wavefile; it is purely Mandarin speech."\\
        - Your response: 0\\
        
        [Example 2 - Inputs]\\
        - "instruction": ""Listen to the audio and assess the clarity and accuracy of pronunciation by considering precise phonology, pronunciation errors, and overall comprehensibility. Use an Arabic numeral from 0 to 10 for your answer.""\\
        - "ground-truth label": 3\\
        - "model output": "The audio is of a man speaking, in a Chinese, saying, "I has to find a different back door". The accuracy of the pronunciation is 90\%."\\
        - Your response: 9\\
        
        [Example 3 - Inputs]\\
        - "instruction": "Please predict the spatial distance(0.5m-4.5m) based on the given audio."\\
        - "ground-truth label": 2.1196093634329904\\
        - "model output": "Based on the given audio, the spatial distance is likely between 0.5m and 4.5m."\\
        - Your response: N/A\\
    
        [Inputs]\\
        - "instruction": \{instruction\}\\
        - "ground-truth label": \{label\}\\
        - "model output": \{model\_output\}\\
        - Your response: 
    }}
    \caption{The prompt used for post-processing the \textbf{regression} tasks inference outputs in the study.}
    \label{fig:regression-prompt}
\end{figure}

\section{Call for tasks}
\label{sec:appendix-call-for-tasks}

\begin{figure}[H]
    \centering
    \includegraphics[width=.9\linewidth]{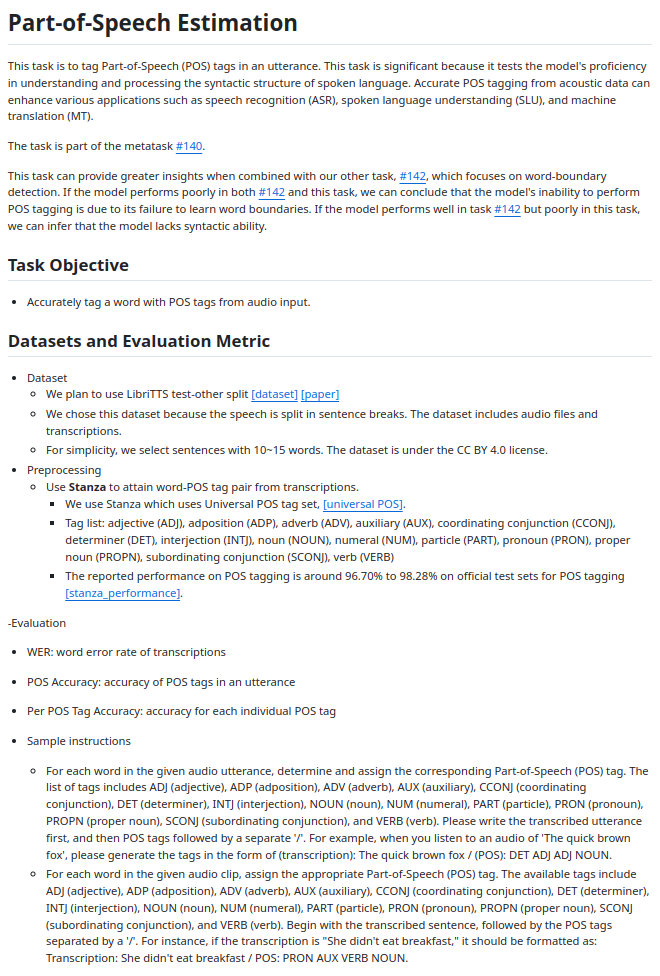}
    \caption{An example of a task proposal from a task contributor.}
    \label{fig:task-proposal-example}
\end{figure}

\begin{figure}[H]
    \centering
    \includegraphics[width=.9\linewidth]{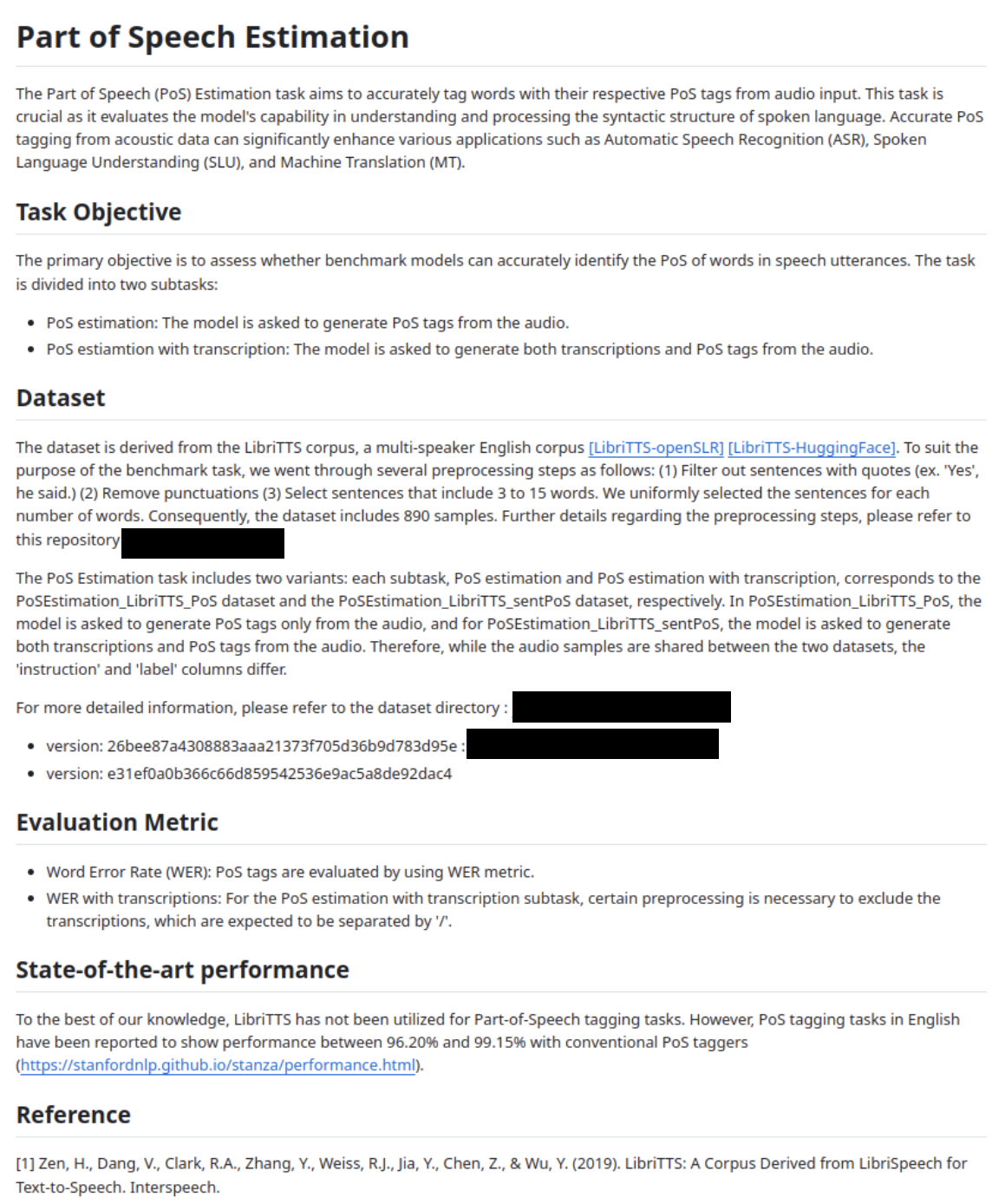}
    \caption{An example of the README file required in the task contribution.}
    \label{fig:task-readme-example}
\end{figure}

\begin{figure}[H]
    \centering
    \includegraphics[width=.7\linewidth]{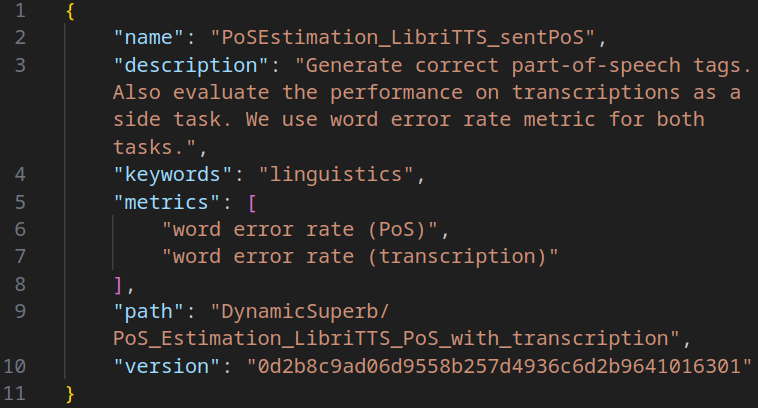}
    \caption{An example of the JSON file required in the task contribution.}
    \label{fig:json-file-example}
\end{figure}

Here we provide material from the call for tasks.
Figure~\ref{fig:task-proposal-example} shows a task proposal that we ask each contributor to initiate.
Task proposers need to provide a high-level description of the task, explain its importance and challenges, and list the datasets they plan to use along with the dataset licenses.
Figure~\ref{fig:task-readme-example} presents the standard README format for each task.
In the README, task proposers include the task introduction, its challenges, the dataset, and the evaluation metric.
Importantly, we also ask them to provide a table of state-of-the-art performance on the task, even if not achieved by universal spoken language models.
These results can serve as a reference for researchers to better understand the task's difficulty and how it has developed so far.
Then, Figure~\ref{fig:json-file-example} shows the JSON format we require to record task information.
For each task, our evaluation pipeline uses this JSON file as input to automatically download data from Huggingface, formatting it for subsequent use.

\section{Audio duration distribution}
\label{sec:audio-duration}
\begin{figure}[H]
    \centering
    \includegraphics[width=.8\linewidth]{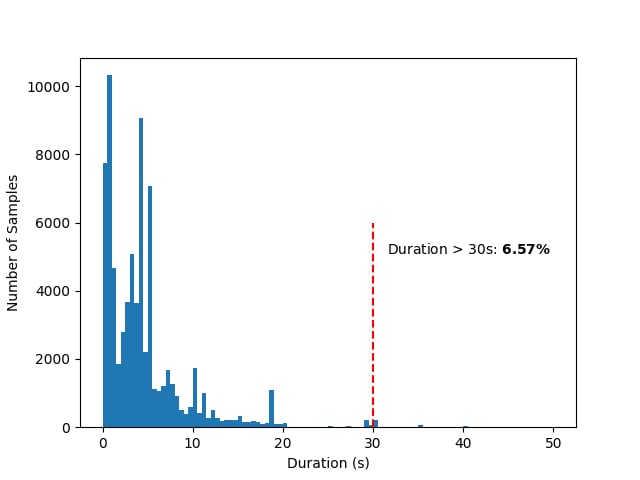}
    \caption{Duration distribution of all audios in Dynamic-SUPERB \textit{Phase-2}.}
    \label{fig:appendix-duration-distribution}
\end{figure}

Figure~\ref{fig:appendix-duration-distribution} shows the distribution of all audio files in Dynamic-SUPERB \textit{Phase-2}.
Only about 6.5\% of the audio files have a duration longer than 30 seconds, which reflects an inherent limitation of models that incorporate Whisper in their framework.

\end{document}